\pdfoutput=1
\documentclass{article}
\PassOptionsToPackage{dvipsnames}{xcolor}
\usepackage{iclr2027_conference,times}
\usepackage[T1]{fontenc}
\usepackage[utf8]{inputenc}
\usepackage{microtype}
\usepackage{booktabs,multirow,makecell,tabularx,array}
\usepackage{xcolor}
\usepackage{colortbl}
\definecolor{topA}{HTML}{4E79A7}\definecolor{topB}{HTML}{B07AA1}\definecolor{topC}{HTML}{F28E2B}
\definecolor{posbg}{HTML}{E3F3EC}\definecolor{negbg}{HTML}{FBE4E4}\definecolor{hdrbg}{HTML}{F1F3F5}
\newcommand{\tA}[1]{\textcolor{topA}{#1}}\newcommand{\tB}[1]{\textcolor{topB}{#1}}\newcommand{\tC}[1]{\textcolor{topC}{#1}}

\newcommand{\grouprow}[2]{\rowcolor{hdrbg}\multicolumn{#1}{@{}l}{\textsc{\footnotesize #2}}\\}
\usepackage{amsmath,amssymb}
\usepackage{pifont}
\usepackage{graphicx}
\graphicspath{{figures/}}
\usepackage{caption}
\PassOptionsToPackage{hyphens}{url}
\usepackage[hidelinks]{hyperref}
\newcommand{\doi}[1]{doi: \href{https://doi.org/#1}{\nolinkurl{#1}}}
\usepackage{cleveref}

\definecolor{passgreen}{HTML}{27AE60}
\definecolor{failred}{HTML}{E74C3C}
\newcommand{\Lpass}{\textcolor{passgreen}{\ding{51}}}
\newcommand{\Lfail}{\textcolor{failred}{\ding{55}}}
\newcommand{\Lpart}{\textcolor{orange}{$\boldsymbol{\sim}$}}
\newcommand{\model}[1]{\texttt{\small #1}}

\newcommand{\lexec}{\textsc{executes}}
\newcommand{\lcorrect}{\textsc{correct}}
\newcommand{\lmethod}{\textsc{method-match}}
\newcommand{\ldirection}{\textsc{direction-match}}

\newcommand{\appinput}[1]{\IfFileExists{#1.tex}{\input{#1}}{\section{[PENDING: \detokenize{#1}]}}}

\title{CausalVerify: End-to-End Verification of\\ Causal Analyses by Language Models}

\author{%
\makebox[\dimexpr\textwidth-2\tabcolsep\relax][c]{%
\parbox[t]{0.44\textwidth}{\centering\normalfont
\textbf{Yonghong Zhang}\thanks{Corresponding author. \texttt{yonghong.zhang@estudiante.uam.es}}\\
Department of Finance\\
Universidad Aut\'onoma de Madrid\\
Madrid, Spain}\hfill
\parbox[t]{0.44\textwidth}{\centering\normalfont
\textbf{Ricardo Correia}\\
Department of Finance\\
Universidad Aut\'onoma de Madrid\\
Madrid, Spain}}\\\noalign{\vskip 1.2em}
\makebox[\dimexpr\textwidth-2\tabcolsep\relax][c]{%
\parbox[t]{0.44\textwidth}{\centering\normalfont
\textbf{Isabel M.~Parra}\\
Department of Finance\\
Universidad Aut\'onoma de Madrid\\
Madrid, Spain}\hfill
\parbox[t]{0.44\textwidth}{\centering\normalfont
\textbf{Yong Xie}\thanks{Corresponding author. \texttt{xieyong.nwpu@gmail.com}}\\
Spanish National Research Council (CSIC)\\
Madrid, Spain}}}
\iclrfinalcopy

\begin{document}
\maketitle

\begin{abstract}
Language models increasingly perform empirical analyses end to end, yet
existing evaluations assess the written explanation or whether generated
code executes, not whether the executed workflow recovers the intended causal
estimand. We introduce \textsc{CausalVerify}, an execution-grounded
benchmark for end-to-end causal analysis that follows a model from
research-context interpretation to estimand recovery. It scores this workflow
at four distinct layers: method recognition, design specification, executable
implementation, and estimand recovery. It combines 259 real-paper contexts, 100 fixed-seed synthetic scenarios with
executable reference estimates, and 23 paper--twin pairs in which a model
commits to a design before seeing the data and its executed analysis is
scored against a canonical estimator on the same realised dataset.

Execution is not correctness. Among 426 model-written workflows that run
without error, 15.5\% fail verification, and a keyword score of the effect
direction stated in the text is a poor proxy for recovery. In a 23-pair,
nine-model paired study, replacing a model's committed design with the reference design and its
execution conventions raises joint recovery of the point estimate and standard
error from 15.0\% to 51.5\%; yet 48.5\% of eligible seeds still fail under
the reference design. The direction replicates on six pairs built afterwards under a frozen construction protocol, although on the two
newest pairs the gain is
confined to models from the family that built the references. Design
specification is consequential but not sufficient: a plausible method and
runnable code do not guarantee recovery, and even supplying the reference design and its conventions leaves substantial
downstream failure.
\textsc{CausalVerify} evaluates the executed workflow rather than its surface
plausibility, and every reported number is recomputed from frozen artifacts
by a single script.
\end{abstract}

\section{Introduction}\label{sec:intro}

\begin{figure*}[t]
\centering
\includegraphics[width=\textwidth]{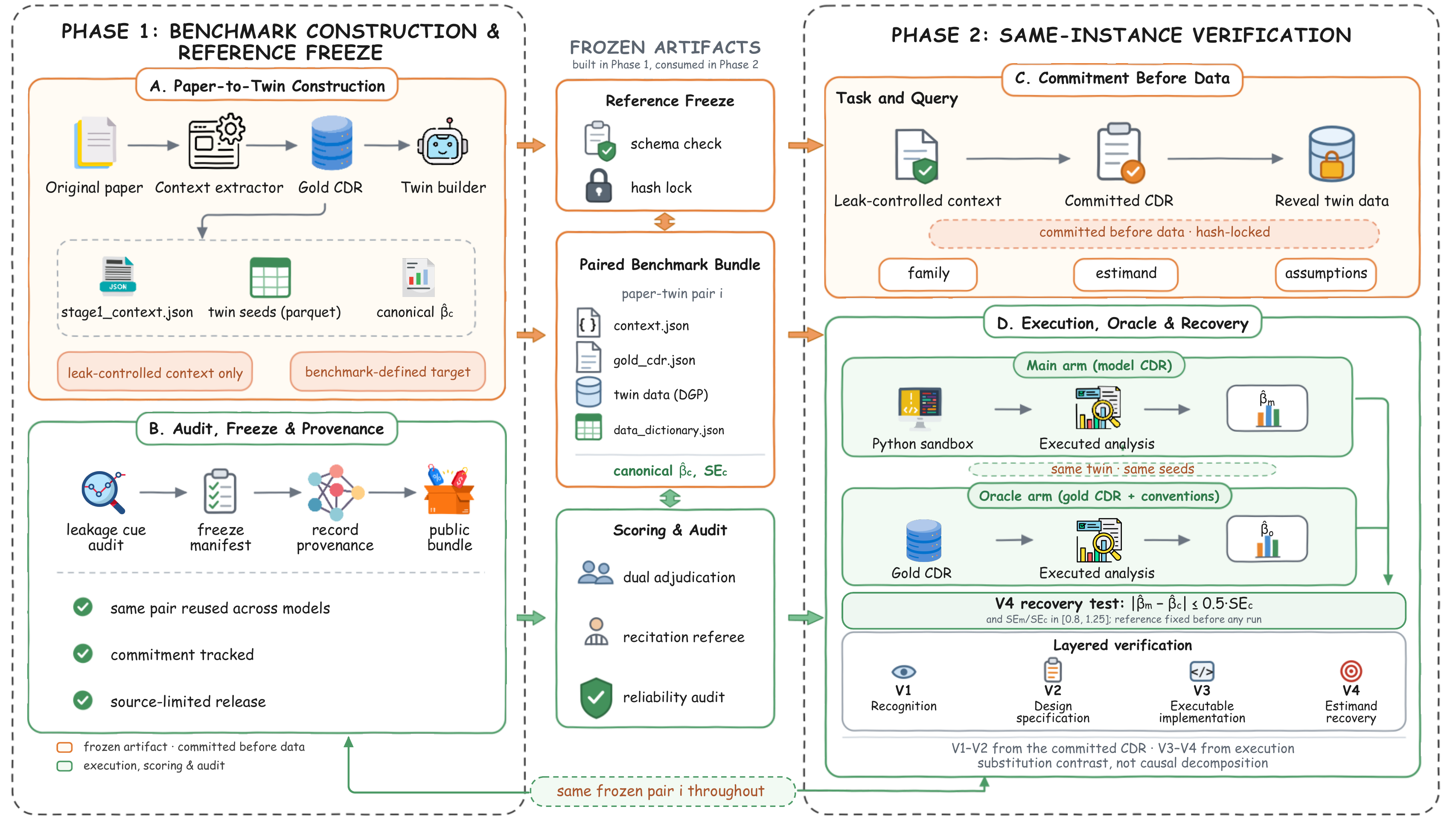}
\caption{\textsc{CausalVerify} separates design commitment from executable
verification. Phase~1 (left) builds a frozen bundle for each paper--twin
pair: a leak-controlled research context, a gold causal-design representation
(CDR), and a synthetic twin whose canonical estimate $\hat\beta_c$ and
standard error are computed by benchmark-side estimators (\textbf{A}), then
audited and hash-frozen (\textbf{B}). Phase~2 (right) evaluates a model
against that bundle only: it commits to its own CDR before any data are
revealed (\textbf{C}), implements it in a fixed sandbox on the twin, and is
scored against the frozen reference (\textbf{D}); the oracle arm reruns the same trajectory with the gold CDR and its pinned
execution conventions substituted. Layers $V_1$--$V_2$ are read
from the commitment, $V_3$--$V_4$ from execution.}
\label{fig:v2-overview}
\end{figure*}

Large language models (LLMs) are moving from commenting on empirical research to
carrying it
out~\citep{korinek2023language,lu2026scientist,majumder2024discoverybench,chen2024scienceagentbench,xie2025autonomouslab}:
in causal analysis, reading an applied setting, committing to an econometric design, writing
estimation code, and reporting a treatment effect. In this workflow, code
that crashes announces itself. The dangerous failure is fluent~\citep{zecevic2023causalparrots}: a response
that names a sensible method, produces code that runs without error, and
reports a confidently signed estimate of the wrong causal quantity. Evaluation practice has largely
kept the instruments of text benchmarking: agreement with reference
labels~\citep{jin2023cladder,jin2024corr2cause,zhou2024causalbench,lee2025econcausal},
plausibility of the described
design~\citep{shi2026intervenebench,sawarni2026causalreasoningbenchmark}, or
whether generated code
executes~\citep{chen2021codex,liu2023evalplus,lai2023ds1000}. Each can tell us
that a model chose a plausible method or produced runnable code; this paper
asks whether these signals predict whether an executed workflow recovers the
intended causal estimand.

A benchmark for this question must be realistic and verifiable at once, and
no single source of tasks is both. Real papers provide authentic research
contexts, but rarely a unique executable reference against which arbitrary
model-written code can be checked~\citep{silberzahn2018many,botviniknezer2020variability,aczel2026analyticalrobustness}.
Synthetic data provide that reference but give up realism. \textsc{CausalVerify} takes realism from real papers and
verifiability from synthetic data, and its paired arm joins the two on the
same item through a synthetic twin matched to a real paper's identification
structure (\cref{fig:v2-overview}). In every executable task, a canonical
estimator fixed by the benchmark defines the target estimate on the same
realised dataset the model analyses, so that ``the code ran'' and ``the
estimate is right'' are scored apart.

The paired arm also makes failures locatable: a model first commits to a
structured causal-design representation (CDR) for a real-paper context,
hashed before any data are revealed in the spirit of a pre-analysis
plan~\citep{casey2012reshaping,olken2015promises}; it then implements that
commitment on the twin, and an oracle arm reruns the same trajectory under the reference design and
its pinned execution conventions. Scoring is layered
throughout: method recognition, design specification, executable
implementation, and estimand recovery are distinct validity targets, scored
separately and never aggregated, because passing one does not guarantee
passing the next (\cref{sec:framework}).

Execution is not correctness: among 426 model-written workflows that execute successfully on
the synthetic scenarios, 15.5\% still fail verification, roughly one in six
analyses that look successful. Execution success remains informative at the
model level, but textual agreement on effect direction is a poor proxy for
estimand recovery (\cref{sec:execution}). In a 23-pair, nine-model paired
study, replacing the model's design with the reference design and its conventions
raises joint recovery of the point estimate and its standard error from 15.0\% to 51.5\%,
a substitution contrast on identical seeds rather than a decomposition of
error. Yet 48.5\% of eligible seeds still fail under the reference design.
Design choice therefore matters substantially, but the reference design
alone does not guarantee recovery (\cref{sec:paired,sec:v3}).

Our contributions are threefold. (1)~An execution-grounded benchmark for
causal-estimation workflows: 259 real-paper contexts, 100 fixed-seed synthetic scenarios, and 23
paper--twin pairs across four quasi-experimental design families; every task begins after the question, data, and setting have been
specified (\cref{sec:benchmark}). (2)~A layered validity framework separating method
recognition, design specification, implementation, and estimand recovery
(\cref{sec:framework}). (3)~Evidence that neither runnable code nor the reference design alone
guarantees recovery of the target estimate, together with an audit of the benchmark's own construction: which parts
were built by language-model agents or scored after results existed, and how
far each moves the numbers (\cref{sec:threats}).

\section{A Layered Verification Framework}\label{sec:framework}

An evaluation that reports a model ``can do causal inference'' asserts
that its score measures the capability named~\citep{raji2021everything}.
Producing
a correct empirical causal result decomposes into
capabilities that are conceptually distinct, separately measurable, and, as
our experiments show, empirically separable (\cref{tab:framework}).

\begin{table}[t]
\centering
\caption{Four validity targets conflated by evaluations of AI-generated
empirical research, and their measurement instruments in
\textsc{CausalVerify}.}
\label{tab:framework}
\small
\begin{tabular}{@{}lll@{}}
\toprule
Layer & Validity target & Instrument \\
\midrule
Method recognition & reference causal-design family & \lmethod\ vs.\ reference labels \\
Design specification & estimand, roles, assumptions & CDR fields vs.\ gold CDR \\
Executable implementation & declared analysis runs & \lexec\ (execution harness) \\
Estimand recovery & target quantity recovered & \lcorrect\ vs.\ canonical estimate \\
\bottomrule
\end{tabular}
\end{table}

Write $V_1, \dots, V_4$ for passing the four layers of \cref{tab:framework}
in order. The framework's central claim is one of \emph{no general
implication}: passing an upstream layer does not guarantee passing the next,
\begin{equation}\label{eq:chain}
V_1 \;\not\Rightarrow\; V_2 \;\not\Rightarrow\; V_3
\;\not\Rightarrow\; V_4 .
\end{equation}
This is weaker than statistical independence (adjacent layers are strongly
correlated: execution ranking tracks recovery at $\tau = 0.81$), and it is an
empirical claim, tested rather than assumed. \Cref{sec:execution} supplies counterexamples
to $V_3 \Rightarrow V_4$ at scale and shows that outer-layer rankings do not
proxy for $V_4$; \cref{sec:paired} supplies same-instance evidence against
$V_2 \Rightarrow V_3$ and $V_2 \Rightarrow V_4$. An evaluation that stops
at $V_1$ measures recognition, one that stops at $V_2$ measures
specification, and one that stops at $V_3$ establishes executability; none of
these endpoints establishes that the target quantity was recovered.

Two design consequences follow. First, criterion validity at the recovery layer requires an \emph{executable
reference target}: a realised dataset on which a
benchmark-fixed canonical estimator defines the target. The canonical
estimator is a benchmark-fixed reference path, not a claim that one analysis
is uniquely correct: the benchmark asks whether the model recovered its stated
target, not whether that target is the only defensible specification
(\cref{sec:benchmark}). Textual reference answers cannot serve as that target~\citep{silberzahn2018many,botviniknezer2020variability,aczel2026analyticalrobustness}: our real-paper labels carry irreducible ambiguity (blinded
audit: direction $\kappa=0.29$). Second, the layers are scored
\emph{separately and never aggregated}. Models may be compared within each
validity target; a composite would re-conflate what the framework separates
and be dominated by whichever layer the task mix happens to weight
(\cref{sec:execution}). The deliverable is a profile, not a ranking.

\section{Related Work}\label{sec:related}

Prior work evaluates important pieces of the LLM causal-analysis workflow~\citep{ma2025causalsurvey,yang2024criticalreview}, but
existing benchmark families terminate at different points of the chain in
\cref{tab:framework}, and rarely at the executed causal estimate.
Causal-reasoning benchmarks test textual
reasoning about direction, counterfactuals, graphs, or
identification~\citep{kiciman2023causal,jin2023cladder,jin2024corr2cause,chen2024clear,chi2024unveiling,zhou2024causalbench,lee2025econcausal}.
Code and science-agent benchmarks test whether programs execute, pass
functional tests, or deliver a final
product~\citep{chen2021codex,lai2023ds1000,jimenez2024swe,zhuo2024bigcodebench,jain2024livecodebench,huang2024mlagentbench,majumder2024discoverybench,chen2024scienceagentbench,anthropic2026biomysterybench}.
Closest are end-to-end causal benchmarks. CauSciBench~\citep{acharya2025causcibench}
grades the final product on curated real, synthetic, and textbook tasks.
\citet{sawarni2026causalreasoningbenchmark} disentangle identification from
estimation on real datasets.
CausalDS~\citep{leban2026causalds} spans Pearl's three rungs on synthetic
structural causal models. InterveneBench~\citep{shi2026intervenebench} scores
agreement between predicted and expert-verified study designs, while
reproduction evaluations~\citep{kohler2026readthepaper,nguyen2026replicatorbench}
use agreement with the original authors' results as the criterion. \textsc{CausalVerify} differs not in task breadth but in its unit and
terminal criterion of verification:
a design committed and hashed before any data are revealed, executed
model-written code, a benchmark-fixed canonical estimate on the same
realised dataset, and an oracle arm that substitutes the reference design and its execution
conventions into the model's own trajectory. Design-layer and execution-layer failures
thus separate on the same instance rather than across task pools. The distinction is not between textual and
executable benchmarks, but between verifying an intermediate artifact and
verifying the target quantity the workflow is meant to produce
(extended comparison in \cref{sec:appendix-related},
\cref{tab:benchmark-comparison}).

\section{Benchmark and Protocol}\label{sec:benchmark}

No single setting provides both ecological realism and executable reference
estimates. Real papers test whether models parse applied research contexts,
but their reference labels are interpretive: cleaned data, runnable specification and counterfactual effect are usually
unavailable (\cref{sec:framework}). Synthetic
data-generating processes (DGPs) give up realism but make coefficient
correctness executable. A paired arm links these two strengths on the same instance, allowing design
interpretation and executed recovery to be evaluated together.

\paragraph{Scope.}\label{sec:scope}
Every task begins after the research question, data description, and
institutional context have been specified. The benchmark evaluates the
downstream portion of empirical practice (design recognition through estimation and reporting) for four
quasi-experimental families~\citep{angrist2009mostly,imbens2015causal}:
difference-in-differences (DID)~\citep{bertrand2004trust,goodmanbacon2021did,roth2023trending}, event studies (ES), instrumental variables
(IV), and regression discontinuity (RDD)~\citep{calonico2014robust}; the paired corpus also includes one
randomized-encouragement design, scored under its IV family~\citep{angrist1996identification}. It does not
evaluate problem formulation, data acquisition, design selection from
ambiguous field constraints, or iterative robustness analysis, and it does not
cover matching, propensity-score, doubly-robust, mediation, or
heterogeneous-effect workflows. Competition-style evaluations that overweight
modeling at the expense of subject-matter reasoning are ``somewhat detached
from the practice of causal inference''~\citep{hernan2019comment}; we claim
nothing outside the stated boundary.

\paragraph{Executable arm: fixed-seed DGP scenarios.}\label{sec:expb}
This arm isolates $V_3$ from $V_4$: it asks separately whether model-written
code runs and whether the executed estimate recovers the benchmark target. It
contains 100 fixed-seed scenarios (30 DID, 24 ES, 24 IV,
22 RDD)~\citep{callaway2021did,sun2021eventstudy,mackinlay1997event,
imbens1994identification,angrist1996identification,
thistlethwaite1960regression,imbens2008rdd,lee2010rdd}. Realised CSV datasets
are fixed before any model evaluation; canonical estimates are computed from
those datasets alone by benchmark-side estimators that never see a model
output (\cref{sec:threats} gives the scorer's history). Models receive
the research question, data description, and a preview of the realised CSV,
then produce runnable R code in one call with a 4{,}096-token output cap
(pipeline in \cref{fig:v1-pipeline}). \lexec\ records whether the code
executes in the fixed environment. \lcorrect\ records whether the extracted treatment-effect
estimate $\hat\beta_m$ matches the canonical estimate $\hat\beta_c$ computed
on the \emph{same realised dataset}:
\begin{equation}\label{eq:tolerance}
|\hat\beta_m - \hat\beta_c| \,/\, |\hat\beta_c| \le \tau,
\qquad \tau = 0.50 \text{ by default.}
\end{equation}
Comparing against the realised-data canonical estimate rather than the ideal
DGP parameter avoids penalizing finite-sample deviation; the estimator's status as a benchmark-fixed
reference path rather than a claim of unique scientific validity is stated in
\cref{sec:framework}. The tolerance is a
permissive reporting default, not an adjudication rule: the model ranking is
identical at every tested tolerance from 10\% to 100\%
(\cref{sec:appendix-tolerance}). A text-direction diagnostic, \ldirection, is
scored on the same scenarios against the known generating sign.

\paragraph{Real-paper contexts as text diagnostics.}\label{sec:expa}
The breadth arm contains 259 archived economics papers, each represented by
reconstructed research-question, data-description, and institutional-context
fields with method names blacklisted. \lmethod\ and \ldirection\ are scored
there against consensus reference labels from four LLM voters, two of which
(GPT-4o, Gemini~2.5 Flash) are also evaluated (\cref{sec:limitations}); an
internal audit, still open, found identification arguments or source results
in more than half of 18 manually reviewed contexts. These labels diagnose the
outer layers only and are never correctness endpoints. A blinded 30-paper
audit on a sample over-representing low-consensus papers agreed with
consensus method labels at $\kappa = 0.61$ and direction labels at
$\kappa = 0.29$, with agreement deteriorating under weak consensus
(\cref{sec:appendix-human}).

\paragraph{Paired bridge: design commitment before data.}\label{sec:bridge}
The paired protocol connects realism and executable verification on the same
item. In Stage~1 a model
reads a real-paper brief and commits to a structured causal-design
representation (CDR: method family, estimand, variable roles, comparison
group, timing, fixed effects, clustering, assumptions); the commitment is
hashed before any data are revealed~\citep{casey2012reshaping,olken2015promises}. In Stage~2 a fresh call hands the model
its own committed CDR together with the data dictionary and column schema of a
matched synthetic twin that preserves the paper's identification structure.
The model writes code implementing its commitment; the script reads the twin
data only inside the sandbox, and estimates and standard errors are scored
against the twin's canonical values under frozen bands. An oracle arm reruns the same Stage-2 trajectory with the benchmark's gold
CDR substituted for the model's commitment;
gold CDRs were drafted and source-checked by language-model agents and
approved by an author (\cref{sec:threats}). The oracle view also fixes the execution conventions the gold
implies (standard-error form, degrees of freedom, sample rules, units). Stage-1 scores are frozen and deviating trajectories remain in the
analysis (frozen protocol in \cref{sec:appendix-protocol}).

\paragraph{Controls, leakage, and scorer audits.}\label{sec:controls}
Because \textsc{CausalVerify} is itself a measurement instrument, its
construction and scoring pipeline is audited separately from model
performance. Separate stages construct contexts, audit leakage, collect responses,
extract coefficients, and aggregate labels, a role separation that reduces
task-writer/respondent/grader circularity; calibration is elicited
separately~\citep{kadavath2022know,tian2023calibration,xiong2024uncertainty}. Templates, effect signs, sample sizes, seeds, realised files, and
canonical estimators are benchmark-fixed, so a model cannot pass \lcorrect\ by
guessing signs or matching phrases. The coefficient-extraction judge is itself
audited. In a blinded 50-cell human validation, on the 44 cells where the
annotator found a treatment effect to compare, numeric agreement was 90.9\%
(40/44) and agreement on the induced pass/fail decision 88.6\% (39/44);
\lcorrect\ is thus an audited instrument, not an oracle. The arms are never aggregated into a single leaderboard;
\cref{tab:controls} in the appendix maps each control to the claim it protects.

\section{Execution Results: Runnable Is Not Right}\label{sec:execution}

The executable arm's two scores, \lexec\ and \lcorrect, separate sharply,
and they do so differently across models.

\begin{figure}[t]
\centering
\includegraphics[width=\textwidth]{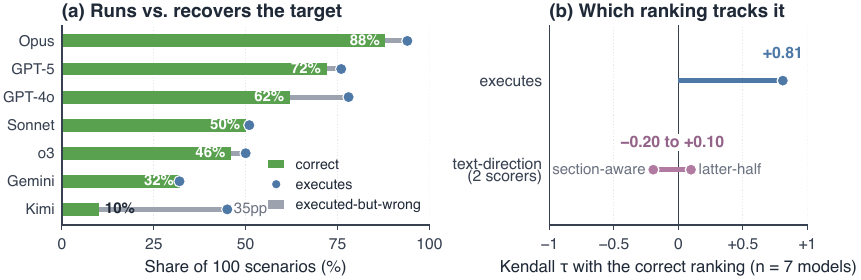}
\caption{Executable arm, headline. \textbf{(a)} Per model, the share of 100
fixed-seed scenarios whose code runs (\lexec, dot) and whose executed estimate
recovers the target within the 50\% band (\lcorrect, bar); the grey segment is
executed-but-wrong. \textbf{(b)} Model-level rank agreement with the
\lcorrect\ ranking: ranking by \lexec\ tracks it (Kendall $\tau = 0.81$); the
two keyword text-direction scorers do not ($\tau \in [-0.20, 0.10]$). Seven
original panel models; open-weight members in \cref{tab:execarm-permodel}.}
\label{fig:exec-headline}
\end{figure}

\paragraph{Execution and recovery are different capabilities.}
Across the seven original panel models, \lcorrect\ pass rates span 10\%--88\%
of 100 scenarios (\cref{fig:exec-headline}a, \cref{tab:execarm-permodel}); the two open-weight panel
members, Qwen3-235B and DeepSeek-V4, reach 34 and 54, with robustness-only
Llama-3.3-70B at 20 (\cref{sec:appendix-openweight}). Of the 426 workflows that execute
successfully in the original panel, 66 (15.5\%; scenario-clustered 95\% CI
$[12.1\%, 19.2\%]$) fail verification: 40 (9.4\%) return an estimate outside
the tolerance band and 26 (6.1\%) yield no extractable target coefficient. We call this union \emph{executed-but-wrong}. Evaluation that stops at ``the
code ran'' classifies all 426 as successes, and the share is not an artifact of a tight band: it is 27.2\% at a 10\% tolerance and still 14.1\% at
100\% (\cref{tab:tolerance-sweep}).

\paragraph{Models fail at different layers.}
Conditional on executing, the panel splits into two regimes. Strong executors
are near-ceiling once code runs (Sonnet 98.0\%, Opus 93.6\% conditional pass),
so their losses concentrate upstream in execution failures. Gemini's 100\% is
not comparable: most of its upstream losses are truncated responses
(\cref{tab:execarm-permodel}).
Weaker executors more often produce workflows that run but fail downstream
verification (Kimi 22.2\% conditional pass: 35 of its
45 executing workflows fail verification). The open-weight families sharpen
the contrast: conditional on execution, Qwen3 fails verification in 2.9\% of
cases and DeepSeek-V4 in 8.5\%, while robustness-only Llama fails in 51.2\%.
These are independently developed families concentrating their failures at
different layers of \cref{tab:framework} (\cref{sec:framework}).

\paragraph{Execution ranking tracks recovery; text-direction ranking does not.}
Model-level rankings by \lexec\ track rankings by \lcorrect\ at Kendall
$\tau = 0.81$ (Spearman $\rho = 0.93$; 1{,}000 scenario-clustered bootstrap
resamples, 95\% CI $[0.62, 0.90]$), and the \lcorrect\ ordering is unchanged at
every tolerance from 10\% to 100\%. Leave-one-model-out panels preserve $\tau \in [0.733, 0.867]$ (leave-Gemini-out
0.867), and removing any one design family leaves the
\lcorrect\ ranking within $\tau \in [0.81, 1.00]$ of the full panel with the
top-two and bottom-two model sets intact (\cref{sec:appendix-execarm}). Execution is therefore informative at the
model level, but not sufficient at the workflow level. A text-level signal
behaves differently. We
score each response's stated effect direction against the known generating
sign with two deterministic keyword scorers, deliberately simple
instruments of the kind a text-only evaluation would use, not validated direction classifiers. Their model rankings correlate with \lcorrect\ at
$\tau \in [-0.20, +0.10]$. At the level of the single workflow, the score
agrees with the true sign for 67.8\% of correct workflows and 60.0\% of
incorrect ones, so it carries little information about whether the computed estimate is
right. We draw no conclusion from any individual model's
text-direction rank, which rests on margins of a few scenarios.

\section{Paired Paper-to-Execution Analysis}\label{sec:paired}

The paired study asks the same question one layer up, on the same item: when
a model commits to a design for a real-paper context and then implements it on
a matched twin, does the commitment carry through to the estimate?
(\cref{fig:v2-overview}C,D; \cref{sec:appendix-protocol}.) The study is reported as a layer profile with pair-clustered uncertainty,
not as a leaderboard.

\paragraph{Pilot and prototype: where chains break.}
A six-pair, one-model pilot (\cref{sec:pilot}) validated the harness: every
committed CDR named the correct design family, so every failure occurred
downstream of $V_1$. An earlier four-task prototype (\cref{sec:prototype})
supplied the motivating case: on its RDD task all nine models executed, and
two vendors' models returned the \emph{same} wrong estimate, 5.075 against a
canonical 1.428 ($3.55\times$), from an \texttt{rdrobust} call on truncated
support, visible only by comparison against the canonical estimate. These
cases motivated the full paired panel by showing that recognition and
execution can both succeed while target recovery fails.

\subsection{The full panel: 23 pairs, nine models}\label{sec:panel}

\begin{figure}[t]
\centering
\includegraphics[width=\textwidth]{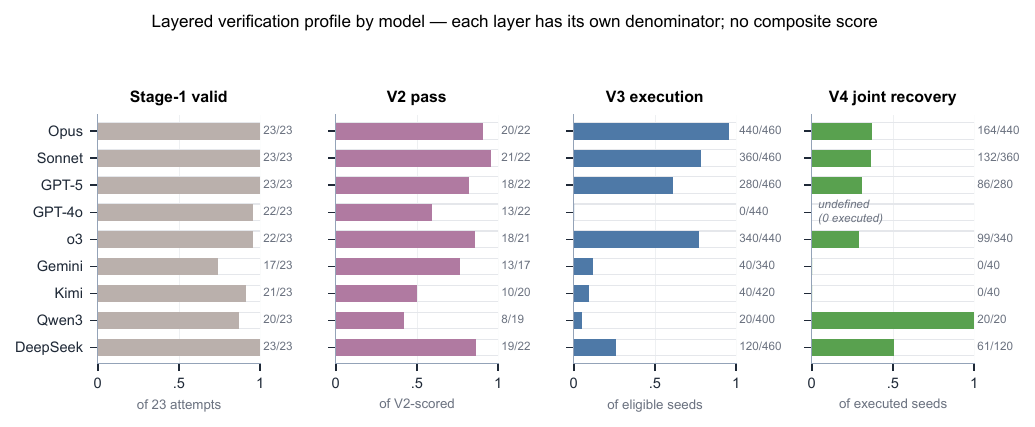}
\caption{Layered verification profile across the nine-model panel. Each
layer is reported against its own denominator (attempts; V2-scored
commitments; eligible seeds; executed seeds) and no composite score is
formed; per-model V2 values are a profile, not a ranking (\cref{sec:threats}).
GPT-4o's recovery layer is \emph{undefined} (0 executed seeds
after a uniform contract violation), which is distinct from the defined
zeros of models that executed and failed to recover.}
\label{fig:panel-cascade}
\end{figure}

The full panel runs all 23 pairs across nine models
(\cref{tab:appendix-pins}; \cref{fig:panel-cascade}) under identical contracts: provider-default sampling,
single-shot with no retry or repair, up to three calls per pair (Stage-1
commitment, main arm, oracle arm; 608 calls in all), an 8{,}192-token output cap, and an execution contract
restricted to numpy, pandas, pyarrow and the Python standard library. Of 207
Stage-1 attempts, 13 produce no schema-valid commitment and are recorded as
not scorable (never as design failures), leaving 194 committed designs. The cap binds: 35 calls stopped at it, accounting for 6 of the 13 invalid
commitments and for 29 Stage-2 calls that executed no seed (per-model
breakdown in \cref{sec:appendix-panel-details}).
Design-specification verdicts were produced by language-model agents working
under the authors' rulings, not by human raters. Adjudication was outcome-blind with sealed model identities, and a
separate reference review checked every flagged gold claim against the
archived source (\cref{sec:appendix-panel-details}). Mismatches
concerning conventions absent from the model-facing context were removed from
scoring as benchmark-underdetermined rather than charged as model error,
uniformly and irrespective of how many models exhibited them. One pair shipped
a defective gold and is excluded layer-scoped. A seeded 20\% re-adjudication of
eligible \textsc{match} verdicts agreed at 97.6\% and a 100\% re-review of
hedge/deferral verdicts at 17/17 (\cref{sec:appendix-panel-details}).

\paragraph{Design layer ($V_2$).} The aggregate $V_2$ pass rate is
descriptive, not a capability estimate: it depends on the adjudication
instrument. Of 187 scored commitments (194 valid
minus the excluded pair's seven), 140 pass
(74.9\%, pair-clustered 95\% CI $[66.1, 82.9]$) and 47 (25.1\%) carry at
least one critical design mismatch attributable to the model on
model-facing evidence: design-family recasts, misreads of explicitly stated
estimands, violations of context-visible sample rules, uninvited
transformations, and internally inconsistent commitments. Per-model pass
rates span 0.42 (Qwen3) to 0.95 (Claude Sonnet~4.6). We rest no claim on its level: on the raw blinded verdicts, before any mismatch was voided, only 23 of 187 commitments
(12.3\%) carry no critical mismatch; 55 (29.4\%) once the inference-convention
field left the scored set; 140 only after the reference review voided 318 mismatches that the
model-facing context did not determine, decisions taken after the verdicts
had been read, uniformly across models.

\paragraph{Conditional downstream execution and recovery.} At the
model$\times$pair unit (seeds are within-trajectory precision, never
independent observations), V2-pass trajectories execute 47.9\% of eligible
seeds against 25.5\% for V2-fail ($+22.3$pp, CI $[4.3, 40.4]$); among
executed seeds, recovery of the canonical estimate is 38.7\% vs.\ 17.1\%
($+21.6$pp, CI $[-7.2, 45.1]$, on twelve executing V2-fail trajectories;
directionally consistent, not resolved). Under a two-way bootstrap over models
and pairs~\citep{efron1994bootstrap} the execution difference's interval also includes zero
($[-13.4, 51.2]$). These are conditional
associations: $V_2$ status is not randomized (\cref{fig:panel-conditional}). Denominators stay separate throughout, and a model with no executed seeds
has undefined recovery rather than zero (GPT-4o: 0/440 after a uniform
contract violation; \cref{fig:panel-cascade}).

\begin{figure}[t]
\centering
\includegraphics[width=0.62\textwidth]{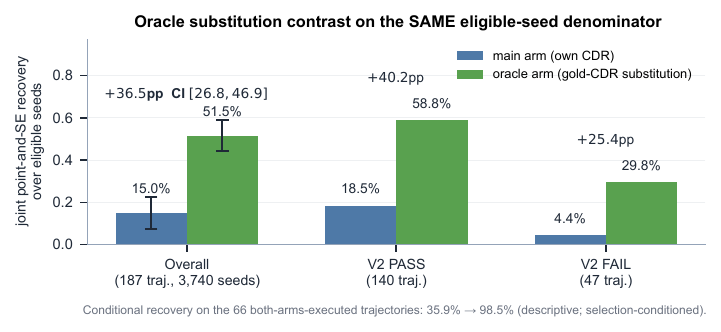}
\caption{Oracle substitution contrast on the common denominator: joint
point-and-SE recovery over \emph{identical eligible seeds}, main arm
(model's own committed design) vs.\ oracle arm (gold-CDR substitution),
overall and by V2 stratum. Error bars are pair-clustered bootstrap 95\%
CIs (22 clusters, $B{=}10{,}000$); cvp\_011 excluded (reference defect).
Conditional recovery on the both-arms-executed subset appears only as a
footnote because it is selection-conditioned.}
\label{fig:panel-oracle}
\end{figure}

\paragraph{Oracle substitution, common denominator.} The oracle arm asks how downstream performance changes when the model's committed design is replaced by the reference design
together with its pinned execution conventions, within the same
model$\times$pair trajectory. The substitution yields three quantities. Execution rises by 12.8pp
(CI $[7.0, 18.8]$). On the 66 trajectories where both arms execute,
conditional recovery rises from 35.9\% to 98.5\% (descriptive;
selection-conditioned). The primary comparison holds the denominator fixed:
joint point-and-SE recovery over identical eligible seeds rises from
15.0\% to 51.5\% ($+36.5$pp, CI $[26.8, 46.9]$), and the contrast is
larger among V2-pass trajectories ($+40.2$pp) than V2-fail ($+25.4$pp).
The substitution effect shows that design information is consequential;
the 48.5\% of eligible seeds that still fail under the gold design show
that it is not sufficient. Correcting the design does not remove the need for downstream
verification. Because the oracle view carries the
gold design \emph{together with} its execution conventions, which the main arm
must supply from its own commitment, the contrast bundles design and
convention information, one more reason it is not a decomposition. The
main arm's Stage-2 prompt carries no execution conventions, so its
standard-error failures cannot be traced to an instruction the model saw;
recovery of the point estimate alone rises from 21.1\% to 53.6\%
($+32.6$pp, CI $[23.0, 42.8]$), so the contrast does not rest on the
standard-error conventions. The
gold designs and oracle conventions were built with Claude-family agents, and
the two Claude models in the panel recover best under them; without those two
models the contrast is 9.3\% to 39.4\% ($+30.1$pp), so it does not depend on
them.
Leave-one-model and leave-one-pair ranges move the substitution contrast by at most $\pm 3.7$pp.

\subsection{Prospective replication on six new pairs}\label{sec:v3}

To test the substitution result prospectively, we built six further pairs
only after the 23-pair panel had been analysed, under a stricter
construction protocol frozen in advance, and ran the same single-shot
contract on them (52 measured cells; one endpoint had been retired by its
provider and was not substituted). Pooled over the six pairs, using each
oracle arm's first run, main-arm joint recovery is 6.1\% of 920 eligible
seeds and oracle recovery 24.1\%: the direction of the substitution contrast
replicates on instances built after the claim was made. Two caveats are
load-bearing. On four of the pairs the oracle arm was re-run after convention
pins were added post hoc (39.8\% instead of 22.2\% on those pairs); only on
the last two were all conventions fixed before any call. On those two pairs the
whole oracle gain (1.1\% to 28.6\%) comes from the two Claude models, which
recover every seed; without them it is 0.5\% to 0.0\%. Most other submissions
fail to execute (9 of 46 calls stopped at the token cap). On these harder
pairs we cannot separate stronger models from models that share conventions
with the agent family that built the references (\cref{sec:appendix-v3}).

\section{Threats We Checked}\label{sec:threats}

This section audits the benchmark as a measurement instrument rather than
the models it evaluates. Every number reported here is recomputed from
released files by the reproduction scripts of \cref{sec:appendix-repro-pins}.

\textbf{Benchmark built with language-model agents.} The 23 gold CDRs, the
synthetic twins, the design-layer adjudication and the reference review were
produced by Claude-family agents under the authors' rulings and batch
approval, not by independent human experts. The two blinded human audits of
\cref{sec:appendix-human} are the only human-rater instruments in this paper.
The paired panel includes two Claude models, so we report no model ranking on
the design layer or the oracle arm. We give the substitution contrast
without them: it survives on the 23 pairs (\cref{sec:panel}), not on
the two newest pairs (\cref{sec:v3}). In the executable arm, the briefs of scenarios
s31--s100 were worded by GPT-4o from benchmark-fixed templates; GPT-4o's gap on them is no larger than other models' (correct on 62.9\% of the briefs it worded and 60.0\% of the
others, against 51.2\% and 46.1\% for the other six models).

\textbf{Scoring choices made after outputs existed.} The executable-arm scorer
changed three times after most outputs had been collected: the reference moved
from the planted parameter to the canonical estimate on the realised data (the
two differ by 10\% or more in 58 of 100 scenarios), regex coefficient
extraction gave way to an LLM extraction judge (audited in
\cref{sec:appendix-human}), and event-study estimates were matched
window-aware. Each step lowered the executed-but-wrong share (56.1\%, 33.1\%,
15.5\%), yet the model ordering by \lcorrect\ is identical under the first and
last scorer ($\tau$ with \lexec: 0.81, 0.71, 0.81). Where canonical and planted
effects agree (42 scenarios), the share is 14.9\% ($\tau = 0.81$);
without event studies, whose window rule is permissive, it is 15.9\% (51/320).
Seven RDD scenarios had their recorded sign corrected after outputs existed,
when a generator bug surfaced (data unchanged); without them the
text-direction correlation is still $\tau = 0.10$.

\textbf{Growth and exclusions.} Scenarios s31--s100 were generated after the
first 30 had been run (executed-but-wrong: 12.2\% on the first 30, 16.7\% on
the rest). The open-weight and robustness-only models were added after the
seven-model results were known. An earlier pre-commitment prompting pilot is
excluded from this paper because its responses were truncated at the token cap
for two models and it was scored against a superseded reference. Boundary experiments are reported in \cref{sec:boundary,sec:proxies}.

\section{Limitations}\label{sec:limitations}

\textbf{Scope.} The benchmark evaluates structured econometric estimation
workflows after the question, data and setting are fixed, not problem
formulation, data acquisition, or design selection in the sense of
\citet{hernan2019comment}. A model could excel here and fail at those stages~\citep{raji2021everything}.
\textbf{Labels and construction.} Real-paper labels come from an LLM consensus
whose pool overlaps the evaluated panel; the blinded audit bounds their
reliability (direction $\kappa = 0.29$). The paired arm's gold CDRs and
adjudication are agent-produced (\cref{sec:threats}). Both are diagnostics or
fixed references, not independent human ground truth.
\textbf{Harness.} Output caps bind for models that spend tokens on hidden
reasoning (55 of Gemini's executable-arm responses; 35 paired calls), so
execution rates partly measure the harness, a scaffold effect documented
across agent benchmarks by \citet{zhang2026confound}. \textbf{Synthetic structure and backend.} Executable reference estimates
require synthetic twins covering four design families, not the space of
applied practice; the executable arm runs R and the paired arm Python, so
cross-arm comparisons carry a backend difference, and magnitudes should not be
read as backend-general.
\textbf{Scale.} The paired study covers 23 pairs and nine mainly closed or
hosted models, with conditional rates whose pair-clustered
recovery-difference interval includes zero. Rank statements are descriptive
over the evaluated panel. Its endpoints are those callable between March
and August 2026 (\cref{tab:appendix-pins}); two have since been retired, and
newer model generations are not evaluated.
\textbf{Instrument.} Scoring depends on canonical specifications, frozen
tolerances and an audited extraction judge. Two twin amendments (a
unit-contract clarification; the repair of one task's Stage-1 context) and the
design-profile instrument are not result-blind and are recorded as such.
\textbf{Contamination.} Source papers are published, so pretraining familiarity
cannot be excluded~\citep{jain2024livecodebench}; identifiers are opaque and per-twin truths sealed.
\Cref{sec:appendix-limitations} expands each point.

\section{Conclusion}\label{sec:conclusion}

\textsc{CausalVerify} treats the executed workflow as the unit of evaluation
for LLM-assisted causal inference and instruments it in the four layers of
\cref{tab:framework}. On the executable arm, 15.5\% of running workflows fail
verification. The paired protocol makes the separation
same-instance: across 23 pairs and nine models, main-arm joint recovery over
eligible seeds is 15.0\%; substituting the gold design and its conventions raises it to 51.5\%,
and 48.5\% of eligible seeds still fail: design specification is
consequential but not sufficient.
\textsc{CausalVerify} measures whether a model-written analysis recovers a benchmark-fixed estimate; its oracle substitution is a measurement
contrast, not a prescription for what should have been done. More generally,
evaluation of quantitative scientific agents should validate the target
quantity itself, and should be built so that others can verify that it did.

\phantomsection\label{mainmatterend}
\subsection*{AI use statement}
Language models were used in three roles: as measurement instruments, as
construction agents, and as coding and writing tools; the paper's claims are
qualified accordingly (\cref{sec:threats}). \emph{As measurement
instruments:} the reference labels of the real-paper arm are a consensus of
LLM voters (four for the method family, three for direction); the executable
arm's coefficient-extraction judge is an LLM (\model{claude-haiku-4-5}, the
same vendor as two evaluated models), audited against blinded human
annotation (\cref{sec:appendix-human}); and the briefs of executable-arm
scenarios s31--s100 were generated by GPT-4o from benchmark-fixed templates.
\emph{As construction agents:} Claude-family models produced the gold
causal-design representations, the synthetic twins, the execution
conventions, the design-layer adjudication verdicts and the reference review
of the paired study, under protocols, rulings and batch approvals issued by
the authors. The two blinded human audits in \cref{sec:appendix-human} are
the only independent human-rater instruments used to validate these
components. \emph{As tools:} AI systems assisted with the evaluation harness,
the scoring and reproduction scripts, figure creation, and manuscript
drafting and revision; the authors reviewed all code and text and are
responsible for every claim. In the categories of the ICLR AI policy, this
covers generating synthetic datasets, designing and giving feedback on
methodology, implementing methods, creating figures, and drafting and editing
the paper. Because two evaluated models belong to the same model family that
constructed the paired references, we also report the substitution contrast
with those models excluded.

\subsection*{Ethics statement}
The benchmark evaluates models on synthetic datasets and on research contexts
reconstructed from archived research papers; it involves no human-subjects data. Paper sourcing carries a contamination surface, handled
by opaque identifiers, sealed reference values, and a deliberate,
versioned release decision rather than default publication. The intended use
is measurement: evidence about where model-written causal analyses fail and
what verification catches. Results should not be read as certifying any
model for unsupervised empirical research, and the benchmark makes no
judgement about which scientific claims a setting licenses: it measures
recovery of benchmark-fixed estimates, nothing more.

\subsection*{Reproducibility statement}
The twin set is frozen under hashed manifests with a dated amendment record
(v1.0--v1.6) in which every post-evidence change states its trigger,
including two amendments that are explicitly not result-blind. Stage-1 inputs
are rendered, frozen artifacts with record-level provenance to their source;
model-facing task identifiers are opaque, with ground truth and identifier
mappings held scorer-side and per-twin truths sealed. Executions run in a
network-isolated container behind a runtime isolation gate that is tested,
not asserted, before any live call; the scorer ships a self-test and a
discrimination table; paid trajectories are archived with committed-design
hashes, per-call usage ledgers, submitted scripts and per-seed numerics under
a hashed evidence freeze, and the design-layer instrument publishes its full
adjudication log. Deferred diagnostic defects are tracked in a public
register and closed by post-hoc audits that may not alter frozen results:
the two closed after unblinding (an infrastructure-sublabel taxonomy pass
and a recitation-zone counterfactual audit) changed zero pass/fail
statuses, and the one presentation-column defect they surfaced is published
as a labeled post-unblind erratum beside the original rather than a silent
correction. A metric crosswalk (\cref{sec:appendix-tolerance}) maps
this paper's names to the released v1 artifacts.

\paragraph{Relation to the v1 release.} The real-paper corpus (259 contexts),
the 100-scenario executable arm and the confidence-elicitation diagnostic were
released earlier as v1, with cached outputs and scoring tables; every
executable-arm number in this paper is recomputed from those frozen files and
is unchanged. New in this paper: the layered verification framework
(\cref{sec:framework}); the paired protocol and its 23-pair panel with the
oracle substitution contrast (\cref{sec:paired}); the prospective replication on
six further pairs (\cref{sec:appendix-v3}); the repair pilot (\cref{sec:appendix-repair}) and the controlled paired-task
repair (\cref{sec:appendix-repair-paired}); the disclosure of construction and scoring choices
(\cref{sec:threats}); and the single-script reproduction harness
(\cref{sec:appendix-repro-pins}). The v1 text-agreement and calibration
results are retained as diagnostics in \cref{sec:proxies}.

\bibliographystyle{iclr2027_conference}
\bibliography{references}

\appendix

\section{Reproduction and Pinned Versions}\label{sec:appendix-repro-pins}

\paragraph{What re-runs.} The release carries a \texttt{reproduce/} directory.
One command verifies the sha256 of every pinned input and then recomputes
each number printed in this paper from the lowest-level released files
(per-(scenario, model) score rows, per-call ledgers with per-seed numerics,
blinded verdict files), comparing it with the printed value and failing on
any difference. No model call is needed. The design-layer freeze, the
unblinded table and the results freeze of the paired study are rebuilt from
the verdicts and rulings, and every bootstrap interval is reproduced with its
recorded seed.

\paragraph{Execution replication.} Re-executing all 700 cached model-written R
scripts of the executable arm in a fresh clone reproduced the \lexec\ flag for
700 of 700 cells and the extracted coefficient for 425 of the 426 executing
cells (the one difference, 1.3\%, comes from an unseeded script and does not
change its verdict). The extraction judge's cached values are not re-run. The
canonical point estimate and standard error of all 460 paired seed-level
references re-derive from the released twin data and estimators.

\paragraph{What does not re-run.} Model generations are cached, not
regenerated: endpoints are provider aliases and would not return the same
text. Three audit counts of the paired study (614 eligible verdicts, 3
disagreements, 17 of 17 hedge re-reviews) exist only in a frozen summary; the
sampled worksheet was not archived. Sealed per-twin truths are withheld by
design and are not needed to recompute any reported number.

\begin{table}[h]
\centering
\caption{Pinned model endpoints. Dates are first and last cached call;
paired-study ledgers carry no timestamps, so their dates are those of the
commits that archived them.}
\label{tab:appendix-pins}
\small
\setlength{\tabcolsep}{4pt}
\resizebox{\textwidth}{!}{%
\begin{tabular}{@{}llll@{}}
\toprule
Arm & Endpoint identifier & Calls & Dates (2026) \\
\midrule
Executable (R; cap 4{,}096) & \model{claude-opus-4-6}, \model{claude-sonnet-4-20250514}, & 100 each & 03-27 to 04-14 \\
 & \model{gpt-4o}, \model{o3}, \model{gemini-2.5-flash}, \model{moonshot-v1-128k} & & \\
 & \model{gpt-5} & 100 & 04-27 to 04-29 \\
 & \model{meta-llama/Llama-3.3-70B-Instruct} (robustness) & 100 & 05-01 \\
 & \model{Qwen/Qwen3-235B-A22B-Instruct-2507}, & 100 each & 07-26 \\
 & \model{deepseek-ai/DeepSeek-V4-Pro} & & \\
Real-paper & the seven executable-arm endpoints above & 259 each & 04-27 to 04-28 \\
Paired (Python; cap 8{,}192) & \model{claude-opus-4-6}, \model{claude-sonnet-4-6}, \model{gpt-5}, & 608 total & 08-21 to 08-22 \\
 & \model{gpt-4o}, \model{o3}, \model{gemini-2.5-flash}, \model{moonshot-v1-128k}, & & \\
 & \model{Qwen/Qwen3-235B-A22B-Instruct-2507}, & & \\
 & \model{deepseek-ai/DeepSeek-V4-Pro} & & \\
\bottomrule
\end{tabular}}
\end{table}

\paragraph{Seeds and scorers.} Bootstrap seeds: 20260507 ($\tau$, $B{=}1{,}000$),
20260424 (executed-but-wrong share, $B{=}2{,}000$), 20260825 (paired study,
$B{=}10{,}000$, 22 pair clusters). Executable-arm verdict: exit code 0 within
60\,s, and relative error below 0.50 against the canonical estimate on the
same realised dataset. Paired verdict: point within $0.5\times$ the canonical
standard error and standard-error ratio in $[0.8, 1.25]$, twin-set freeze
v1.6, 20 committed seeds per pair.

\section{Tolerance Sweep and Metric Crosswalk}\label{sec:appendix-tolerance}

\begin{figure}[t]
\centering
\includegraphics[width=0.58\textwidth]{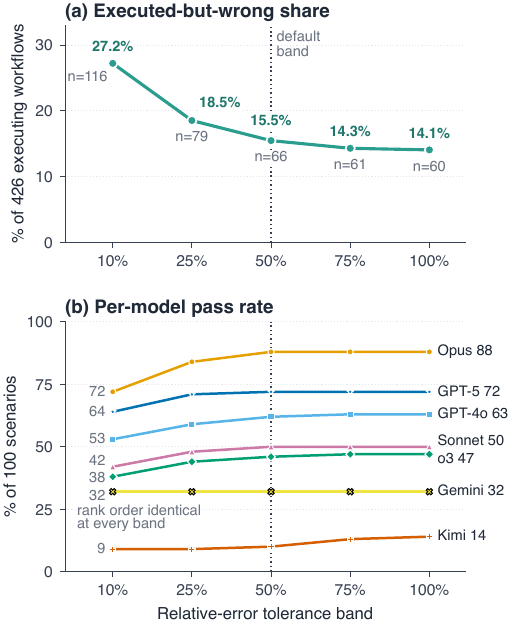}
\caption{Tolerance sweep, drawn from \cref{tab:tolerance-sweep}. \textbf{(a)}
Executed-but-wrong share of the 426 executing workflows as the relative-error
band widens from 10\% to 100\%; the default 50\% band is the dotted line.
\textbf{(b)} Per-model \lcorrect\ rates at each band; the rank order is
identical at every band.}
\label{fig:tolerance-sweep}
\end{figure}

The binary \lcorrect\ endpoint uses a broad default relative-error tolerance
of 50\%. To check that the qualitative conclusions do not depend on this
cutoff, primary-panel pass rates are recomputed at 10\%, 25\%, 50\%, 75\%
and 100\%; the sweep is a robustness diagnostic and the frozen headline
results keep the 50\% default. \Cref{tab:tolerance-sweep} and \cref{fig:tolerance-sweep} report
all-scenario pass rates, so non-executing outputs remain failures. Rankings
are essentially tolerance-invariant. The final row aggregates the
executed-but-wrong phenomenon across the panel:
$P(\text{wrong}\mid\text{executable})$ is 27.2\% at the strict 10\%
tolerance and still 15.5\% and 14.1\% at 50\% and 100\%. The phenomenon
is not an artefact of a tight band.

\begin{table}[h]
\centering
\caption{Tolerance-sweep diagnostic for the executable arm. Each cell is the
pass rate out of 100 scenarios; parentheses give the model rank at that
tolerance. The 50\% column is the frozen headline threshold. The final row
pools the panel: the share of \emph{executing} workflows whose estimate
still falls outside the band. Top three per column in \tA{blue}, \tB{purple},
\tC{orange}; the pooled row is bold and uncoloured.}
\label{tab:tolerance-sweep}
\small
\renewcommand{\arraystretch}{1.08}
\begin{tabular}{lccccc}
\toprule
\textbf{Model} & \textbf{10\%} & \textbf{25\%} & \textbf{50\%} &
\textbf{75\%} & \textbf{100\%} \\
\midrule
\grouprow{6}{Per model}
Opus & \tA{72\% (1)} & \tA{84\% (1)} & \tA{88\% (1)} & \tA{88\% (1)} & \tA{88\% (1)} \\
GPT-5 & \tB{64\% (2)} & \tB{71\% (2)} & \tB{72\% (2)} & \tB{72\% (2)} & \tB{72\% (2)} \\
GPT-4o & \tC{53\% (3)} & \tC{59\% (3)} & \tC{62\% (3)} & \tC{63\% (3)} & \tC{63\% (3)} \\
Sonnet & 42\% (4) & 48\% (4) & 50\% (4) & 50\% (4) & 50\% (4) \\
o3 & 38\% (5) & 44\% (5) & 46\% (5) & 47\% (5) & 47\% (5) \\
Gemini & 32\% (6) & 32\% (6) & 32\% (6) & 32\% (6) & 32\% (6) \\
Kimi & 9\% (7) & 9\% (7) & 10\% (7) & 13\% (7) & 14\% (7) \\
\midrule
\grouprow{6}{Pooled over the panel}
{\bfseries\boldmath $P(\text{wrong}\mid\text{exec.})$} & \textbf{27.2\%} & \textbf{18.5\%} & \textbf{15.5\%} & \textbf{14.3\%} & \textbf{14.1\%} \\
\bottomrule
\end{tabular}
\end{table}

\Cref{tab:crosswalk} fixes the correspondence between this paper's semantic
metric names and the L-codes used by the released v1 artifacts and scoring
scripts, so results can be cross-checked against the frozen records without a
translation step. Appendix text carried over from the v1 release refers to the
real-paper arm as Exp~A and to the executable arm as Exp~B.

\begin{table}[h]
\centering
\caption{Metric crosswalk: semantic names used in this paper versus the
L-codes of the released v1 artifacts.}
\label{tab:crosswalk}
\small
\begin{tabular}{lll}
\toprule
\textbf{This paper} & \textbf{v1 code} & \textbf{What it measures} \\
\midrule
\lmethod & L3 & method-family agreement with reference labels \\
\ldirection & L4 & sign/direction agreement with reference labels \\
\lexec & L2b & code executes in the fixed environment \\
\lcorrect & L2b$^{+}$ & executed estimate inside the frozen band \\
\bottomrule
\end{tabular}
\end{table}

\section{Executable-Arm Details}\label{sec:appendix-execarm}

\begin{figure}[h]
\centering
\includegraphics[width=\textwidth]{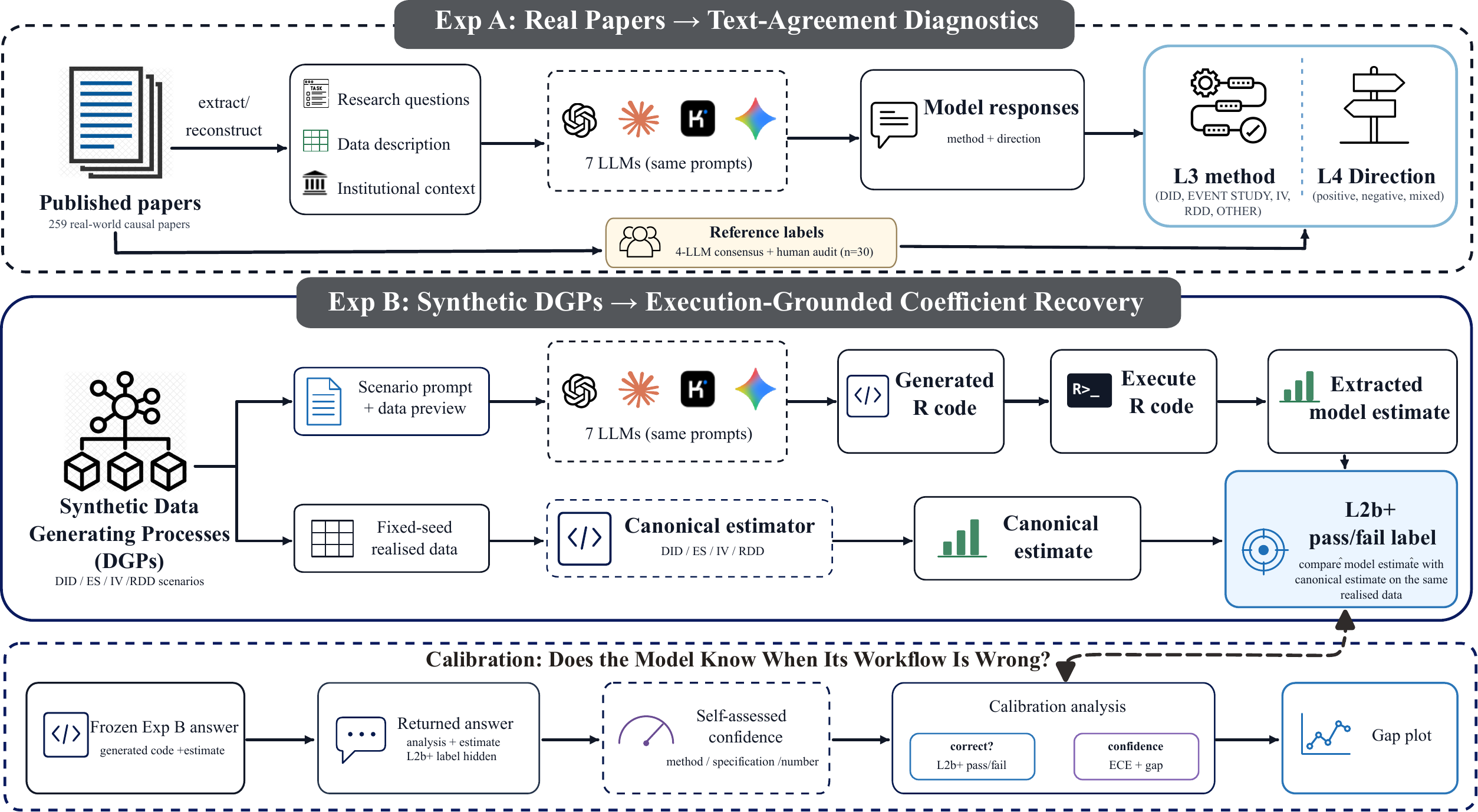}
\caption{The v1 evaluation pipelines as released: the real-paper arm (top;
259 contexts, reference labels from a four-LLM consensus with a blinded human
audit), the executable arm (middle; fixed-seed synthetic scenarios, model-written
R code executed and compared with the canonical estimate on the same realised
data) and the confidence-elicitation diagnostic (bottom). Labels use the v1
vocabulary (L3, L4, L2b$^{+}$); \cref{tab:crosswalk} maps them to this paper's
names. The paired protocol of \cref{sec:paired} is drawn in
\cref{fig:v2-overview}.}
\label{fig:v1-pipeline}
\end{figure}

This appendix reports supporting diagnostics for the executable arm
(\cref{sec:expb}). Every table is derived from the frozen v1 scoring file for
the seven-model v1 panel (700 model--scenario cells) and none replaces the
frozen headline rates of \cref{sec:execution}. Two instruments appear
throughout: \lexec\ (legacy code L2b), whether the model-written code runs,
and \lcorrect\ (legacy code L2b$^{+}$), whether the extracted estimate falls
inside the band of \cref{eq:tolerance} around the canonical estimate.

\begin{table}[t]
\centering
\caption{Executable arm, per model (100 scenarios each). \lexec: code runs;
\lcorrect: executed estimate inside the 50\% band; last two columns: how
executed-but-wrong workflows fail. The two open-weight members were run later
(\cref{sec:appendix-openweight}). ``Sonnet'' here
is \model{claude-sonnet-4-20250514}; the paired study uses
\model{claude-sonnet-4-6}. Gemini's row is depressed by the harness: 55 of its
100 responses stopped at the 4{,}096-token output cap and 43 of those contain no
code block (on its 45 complete responses it executes and recovers 25).
Top three per column across the nine model rows in \tA{blue}, \tB{purple},
\tC{orange}; for the two failure-count columns the three \emph{lowest} are
coloured, and tied values share a tier.}
\label{tab:execarm-permodel}
\small
\setlength{\tabcolsep}{5pt}
\begin{tabular}{@{}lccccc@{}}
\toprule
 & & & \lcorrect\ given & \multicolumn{2}{c}{executed-but-wrong} \\
\cmidrule(l){5-6}
Model & \lexec & \lcorrect & \lexec\ (\%) & outside band & no coefficient \\
\midrule
\grouprow{6}{original panel (seven)}
Opus & \tA{94} & \tA{88} & 93.6 & 6 & \tA{0} \\
GPT-5 & \tC{76} & \tB{72} & 94.7 & 4 & \tA{0} \\
GPT-4o & \tB{78} & \tC{62} & 79.5 & 5 & 11 \\
Sonnet & 51 & 50 & \tB{98.0} & \tB{1} & \tA{0} \\
o3 & 50 & 46 & 92.0 & 3 & 1 \\
Gemini & 32 & 32 & \tA{100.0} & \tA{0} & \tA{0} \\
Kimi & 45 & 10 & 22.2 & 21 & 14 \\
\midrule
\textbf{Pooled (seven)} & \textbf{426} & \textbf{360} & \textbf{84.5} & \textbf{40 (9.4\%)} & \textbf{26 (6.1\%)} \\
\midrule
\grouprow{6}{open-weight, run later}
Qwen3-235B & 35 & 34 & \tC{97.1} & \tB{1} & \tA{0} \\
DeepSeek-V4 & 59 & 54 & 91.5 & 5 & \tA{0} \\
\bottomrule
\end{tabular}
\end{table}

\subsection{Canonical estimators per design family}
\label{sec:appendix-execarm-canonical}

\lcorrect\ compares the model-reported estimate to a fixed realised-data
reference estimate. The canonical estimator is a benchmark-defined reference
path for executable evaluation, not a claim that only one empirical analysis
is scientifically defensible. It is also distinct from the structural DGP
parameter used to simulate the data. \Cref{tab:appendix-execarm-canonical}
lists, per family, the reference estimate, the model-reported coefficients
accepted as addressing it, and the coefficients most often reported instead.

\begin{table}[h]
\centering
\caption{Benchmark-defined canonical estimators used for \lcorrect\ scoring
on the executable arm.}
\label{tab:appendix-execarm-canonical}
\small
\renewcommand{\arraystretch}{1.10}
\setlength{\tabcolsep}{4pt}
\begin{tabularx}{\linewidth}{@{}l
  >{\raggedright\arraybackslash}X
  >{\raggedright\arraybackslash}X
  >{\raggedright\arraybackslash}X@{}}
\toprule
Family & Canonical reference estimate & Accepted model-reported coefficient &
Common non-target coefficients \\
\midrule
DID & TWFE coefficient on \texttt{treat\_x\_post} after unit and period
demeaning & \texttt{treat\_x\_post}, \texttt{treated:post}, or equivalent
treatment-by-post effect & \texttt{treated} main effect, \texttt{post} main
effect, placebo or pretrend coefficient \\
\addlinespace
ES & Market-model coefficient on \texttt{event\_day >= 0}; the final
ES-aware scorer accepts frozen integer post-window conversions to this
per-day scale & Explicit post-event AAR/CAR that can be deterministically
converted to the canonical scale & Pre-event coefficient, arbitrary
non-frozen window, diagnostic or placebo return \\
\addlinespace
IV & Realised-data 2SLS coefficient on \texttt{x}, instrumented by
\texttt{z}, with controls exogenous when present & Second-stage coefficient
on \texttt{x} or the named endogenous variable & First-stage coefficient on
\texttt{z}, reduced form, or control coefficient \\
\addlinespace
RDD & Local-linear cutoff effect on the realised data: coefficient on
\texttt{treated} near cutoff 0.5 with running-variable slope and interaction
& Treatment, above-cutoff, or discontinuity coefficient consistent with
benchmark direction & Density test, bandwidth statistic, unrelated global
polynomial trend, placebo cutoff estimate \\
\bottomrule
\end{tabularx}
\end{table}

\paragraph{Scope of each reference.} Each reference is deliberately narrow.
The DID target is a clean benchmark DID template, not a full benchmark of
staggered or heterogeneous-treatment DID estimators~\citep{callaway2021did,goodmanbacon2021did,sun2021eventstudy}. The event-study target
is a fixed benchmark window mapping, not a universal event-study estimand~\citep{mackinlay1997event}.
The IV target tests implementation of the IV estimand~\citep{imbens1994identification,angrist1996identification}; it does not validate
the exclusion restriction or weak-instrument diagnostics. The RDD target is a
fixed benchmark RDD reference, not an exhaustive comparison of bandwidth or
inference choices~\citep{imbens2008rdd,lee2010rdd,calonico2014robust}.

Because \lcorrect\ requires code execution, the raw association between
\lexec\ and \lcorrect\ is partly structural.
\Cref{tab:appendix-execarm-conditional} therefore reports the conditional
metric $P(\text{\lcorrect} \mid \text{\lexec})$, which asks whether executed
code computes the right coefficient.

\begin{table}[h]
\centering
\caption{Conditional correctness among executed outputs (counts out of 100
scenarios per model). ``Wrong after exec.'' is the \lexec\ count minus the
\lcorrect\ count; the pooled row sums the seven models and is the source of
the 15.5\% (66/426) executed-but-wrong share in \cref{sec:execution}.
Top three per column in \tA{blue}, \tB{purple}, \tC{orange} for the columns
where higher is better (\lexec, \lcorrect, and the conditional rate); the
pooled row is left uncoloured.}
\label{tab:appendix-execarm-conditional}
\small
\renewcommand{\arraystretch}{1.08}
\begin{tabular}{lrrrrr}
\toprule
Model & \lexec & \lcorrect & Wrong after exec. &
$P(\text{\lcorrect} \mid \text{\lexec})$ & Gap \\
\midrule
\grouprow{6}{Per model (100 scenarios each)}
Opus & \tA{94} & \tA{88} & 6 & 93.6\% & 6 pp \\
GPT-5 & \tC{76} & \tB{72} & 4 & \tC{94.7\%} & 4 pp \\
GPT-4o & \tB{78} & \tC{62} & 16 & 79.5\% & 16 pp \\
Sonnet & 51 & 50 & 1 & \tB{98.0\%} & 1 pp \\
o3 & 50 & 46 & 4 & 92.0\% & 4 pp \\
Gemini & 32 & 32 & 0 & \tA{100.0\%} & 0 pp \\
Kimi & 45 & 10 & 35 & 22.2\% & 35 pp \\
\midrule
Pooled & 426 & 360 & 66 & 84.5\% & -- \\
\bottomrule
\end{tabular}
\end{table}

\subsection{Failure-taxonomy diagnostic}
\label{sec:appendix-execarm-taxonomy}

We also audit the cells that fail \lcorrect\ with a conservative rule-based
taxonomy. The taxonomy is diagnostic rather than a scoring replacement: it
distinguishes missing or unparseable code, execution failures,
executed-but-unreported effects, wrong target coefficients, event-window or
scale mismatches, IV stage errors, and RDD cutoff or sign-convention issues.
Ambiguous cases are left uncategorized rather than forced into a narrow
label. In \cref{tab:appendix-execarm-taxonomy} the first two columns (230
execution failures plus 44 no-code cells) are the 274 cells that fail
\lexec, and the remaining six columns sum to the 66 executed-but-wrong cells
of \cref{tab:appendix-execarm-conditional}.

\begin{table}[h]
\centering
\caption{Failure-taxonomy diagnostic on the executable arm, per v1-panel
model. Cells show counts and (share of that model's cells failing
\lcorrect). Strong models execute correctly when they execute (Opus, Sonnet,
Gemini conditional pass rates $\ge 93\%$; see
\cref{tab:appendix-execarm-conditional}), while weaker models more often
produce executed-but-wrong code (Kimi 22.2\%, GPT-4o 79.5\% conditional
pass). Per-model rows sum to 340 failing cells, matching the aggregate
decomposition (Exec fail 230, No code 44, Wrong coef 26, Coef-not-reported
15, Window/scale 11, IV stage 5, RDD 3, Other 6). Bold marks the largest
failure category in each row.}
\label{tab:appendix-execarm-taxonomy}
\small
\setlength{\tabcolsep}{0pt}
\renewcommand{\arraystretch}{1.05}
\begin{tabular*}{\linewidth}{@{\extracolsep{\fill}}lrrrrrrrrr@{}}
\toprule
Model & $n$ & Exec fail & No code & Wrong coef & Coef N/R & Win/scale &
IV stage & RDD & Other \\
\midrule
Opus   & 12 & \textbf{5 (41.7\%)}  & 1 (8.3\%)   & 2 (16.7\%)  & 0           & 3 (25.0\%) & 0          & 0          & 1 (8.3\%) \\
Sonnet & 50 & \textbf{49 (98.0\%)} & 0           & 1 (2.0\%)   & 0           & 0          & 0          & 0          & 0 \\
GPT-4o & 38 & \textbf{22 (57.9\%)} & 0           & 3 (7.9\%)   & 1 (2.6\%)   & 3 (7.9\%)  & 4 (10.5\%) & 0          & 5 (13.2\%) \\
o3     & 54 & \textbf{50 (92.6\%)} & 0           & 2 (3.7\%)   & 1 (1.9\%)   & 1 (1.9\%)  & 0          & 0          & 0 \\
Kimi   & 90 & \textbf{55 (61.1\%)} & 0           & 18 (20.0\%) & 13 (14.4\%) & 3 (3.3\%)  & 1 (1.1\%)  & 0          & 0 \\
Gemini & 68 & 25 (36.8\%) & \textbf{43 (63.2\%)} & 0           & 0           & 0          & 0          & 0          & 0 \\
GPT-5  & 28 & \textbf{24 (85.7\%)} & 0           & 0           & 0           & 1 (3.6\%)  & 0          & 3 (10.7\%) & 0 \\
\bottomrule
\end{tabular*}
\end{table}

\subsection{Rank-stability diagnostic}
\label{sec:appendix-execarm-rank}

The cross-layer rank correlations of \cref{sec:execution} are descriptive
model-level diagnostics over the seven v1-panel models rather than
population-level estimates. To assess panel sensitivity, Kendall $\tau$ and
Spearman $\rho$ between the \lexec\ ranking and the \lcorrect\ ranking are
recomputed after leaving out each \emph{model} in turn. The
leave-one-model-out Kendall range is $0.733$--$0.867$ and the Spearman range
is $0.886$--$0.943$ (\cref{tab:appendix-execarm-rank}). Adding
Llama-3.3-70B-Instruct as a robustness-only eighth model gives Kendall
$\tau=0.714$ and Spearman $\rho=0.881$, while the ranked v1 panel remains
the seven models.

\begin{table}[h]
\centering
\caption{Rank stability of the model-level correlation between the \lexec\
ranking and the \lcorrect\ ranking on the executable arm. The Llama row is
robustness-only and never enters the ranking. Bold marks the headline
full-panel row, with its coefficients in \tA{blue}.}
\label{tab:appendix-execarm-rank}
\small
\setlength{\tabcolsep}{5pt}
\renewcommand{\arraystretch}{1.08}
\begin{tabular}{lccc}
\toprule
Diagnostic & $N$ models & Kendall $\tau$ & Spearman $\rho$ \\
\midrule
\grouprow{4}{Seven-model v1 panel (ranked)}
\textbf{Seven-model baseline} & \textbf{7} & \tA{\textbf{0.810}} & \tA{\textbf{0.929}} \\
Scenario-clustered bootstrap 95\% CI (1{,}000 resamples) & 7 &
{\scriptsize $[0.62, 0.90]$} & {\scriptsize $[0.75, 0.96]$} \\
Leave-one-model-out range & 6 each & 0.733--0.867 & 0.886--0.943 \\
\grouprow{4}{Eighth model added (robustness-only, not ranked)}
Seven plus Llama (robustness-only) & 8 & 0.714 & 0.881 \\
\bottomrule
\end{tabular}
\end{table}

\paragraph{Leave-one-design-out.} A second diagnostic, derived
deterministically from the same frozen scoring file after the v1 release,
removes one design \emph{family} at a time and re-ranks the seven models by
\lcorrect\ on the reduced panel at the default 50\% tolerance. The statistic
is Kendall's $\tau_a$ between the \lcorrect\ ranking on the \emph{full}
panel and the \lcorrect\ ranking on the \emph{reduced} panel (seven models,
21 pairs; a tie counts as neither concordant nor discordant), so it measures
how stable the recovery ranking itself is to the family mix. It is a different quantity
from the cross-layer $\tau(\lexec, \lcorrect)$ of
\cref{tab:appendix-execarm-rank}, and the two sets of values are not
comparable: the DID-excluded value coincides numerically with the
cross-layer baseline (both $17/21 = 0.810$) but correlates a different pair
of rankings. In \cref{tab:appendix-execarm-lodo}, $\tau_a$ is 0.810, 0.905,
0.952 and 1.000; the top-two set \{Opus, GPT-5\} and the bottom-two set
\{Gemini, Kimi\} are intact in all four reduced panels, and every movement
is confined to ranks 3--5.

\begin{table}[h]
\centering
\caption{Leave-one-design-out diagnostic on the executable arm (seven-model
v1 panel, 50\% tolerance). $\tau_a$ compares the reduced-panel \lcorrect\
ranking with the full-panel \lcorrect\ ranking; ranks 1--2 (Opus, GPT-5) and
6--7 (Gemini, Kimi) never change, so only ranks 3--5 are shown. ``Wrong /
exec.'' pools the seven models: executing workflows whose estimate falls
outside the band, over executing workflows. Top three per column in
\tA{blue}, \tB{purple}, \tC{orange} (highest $\tau_a$; lowest share); the
full-panel reference row is in bold and is not ranked.}
\label{tab:appendix-execarm-lodo}
\small
\renewcommand{\arraystretch}{1.08}
\begin{tabular}{llcccl}
\toprule
Panel & Scenarios & $\tau_a$ vs.\ full & Wrong / exec. & Share &
Ranks 3--5 by \lcorrect \\
\midrule
\grouprow{6}{Reduced panels (one family excluded)}
Excluding DID & 70 & 0.810 & 46/285 & 16.1\% & o3 $>$ GPT-4o $>$ Sonnet \\
Excluding ES & 76 & \tC{0.905} & 51/320 & 15.9\% & GPT-4o $>$ o3 $>$ Sonnet \\
Excluding IV & 76 & \tB{0.952} & 51/317 & 16.1\% & GPT-4o $=$ Sonnet $>$ o3$^{\dagger}$ \\
Excluding RDD & 78 & \tA{1.000} & 50/356 & \tB{14.0\%} & GPT-4o $>$ Sonnet $>$ o3 \\
\grouprow{6}{Single-family panels}
DID only & 30 & n/a & 20/141 & \tC{14.2\%} & n/a \\
ES only & 24 & n/a & 15/106 & \tC{14.2\%} & n/a \\
IV only & 24 & n/a & 15/109 & \tA{13.8\%} & n/a \\
RDD only & 22 & n/a & 16/70 & 22.9\% & n/a \\
\midrule
\textbf{Full panel} & \textbf{100} & n/a & \textbf{66/426} & \textbf{15.5\%} & \textbf{GPT-4o $>$ Sonnet $>$ o3} \\
\bottomrule
\end{tabular}

\raggedright\footnotesize
$^{\dagger}$Exact tie, not a reversal: with the IV scenarios removed GPT-4o
and Sonnet both pass 47 of 76, which is why $\tau_a = 20/21$ rather than 1.
\end{table}

The executed-but-wrong share is equally insensitive to the family mix:
14.0\%--16.1\% in every reduced panel against 15.5\% on the full panel, and
13.8\%--14.2\% within DID, ES and IV, with RDD highest (22.9\%) on the
smallest executing base. Across tolerances the same pooled share is 116/426
(27.2\%), 79/426 (18.5\%), 66/426 (15.5\%), 61/426 (14.3\%) and 60/426
(14.1\%) at 10\%, 25\%, 50\%, 75\% and 100\%, completing the three values
shown in the final row of \cref{tab:tolerance-sweep}.

\section{Paired-Analysis Protocol}\label{sec:appendix-protocol}

This appendix records the protocol behind \cref{sec:paired}. Its normative
documents are among the 17 set-wide artifacts hashed by the twin-set freeze;
the panel ran under freeze v1.6 throughout.

\paragraph{Eligibility and gold annotation.}\label{sec:appendix-protocol-eligibility}
Selection rules were frozen before any candidate was ranked~\citep{casey2012reshaping,olken2015promises}; no model output
or benchmark score may be consulted. A paper is eligible only if its
bibliographic identity is externally verified, its inputs are legally
reproducible, and a counterfactual perturbation can be built without changing
the design class: its ``operative identification strategy must belong to one
of the four preregistered design families, and the synthetic twin must
preserve that strategy without reinterpreting the study into another design
class.'' An operative-design re-audit froze a shortlist of 23 (DID~6, IV~7,
RDD~7, event study~3). The gold CDR describes the design, not the authors'
code, with one scored primary estimand per task, chosen where the source
designates none by a frozen hierarchy in which significance, magnitude and
sign are never criteria; golds were drafted and source-checked by
language-model agents, approved in batches by an author, and hashed before
any twin is built.

\paragraph{Twin construction, seeds and sealed truth.}
The CDR is the only interface between paper and twin: the generator consumes
the design payload and twin requirements and nothing else. A twin preserves
the full design structure (instrument and first stage, adoption timing,
running variable, compliance types); design-class flips are prohibited. Paper
identity is conceded; protection is value-level. Each planted coefficient
draws a regime (strong positive, strong negative, near null) by a sealed
seed, block-balanced within family, and observed sign, significance and null
status are not permitted inputs. The planted magnitude is $|\tau^{*}| = t^{*}
\times SE_0$, with $SE_0$ the null-scaffold canonical SE and $t^{*} \sim
U(2.5, 6)$ (strong) or $U(0, 1)$ (near null); scales carry a per-twin factor
from $U(0.25, 4)$, and $\tau^{*}$ is redrawn if within $2\times$ the scoring
band of a comparable published value. Acceptance (recovery, coverage, SE
calibration) uses at least 300 internal seeds per twin under pre-registered
bands; \emph{20 seeds per pair} are committed for evaluation. Each seed
file's canonical estimate and SE sit in a sealed truth file, committed by
hash and never exposed, and seed $k$ is scored against seed $k$'s values:
match-to-canonical, not match-to-$\tau^{*}$.

\begin{sloppypar}
\paragraph{CDR schema.}\label{sec:appendix-protocol-schema}
A CDR is a JSON object with exactly four top-level fields: \texttt{family}
(\texttt{DID}, \texttt{IV}, \texttt{RDD}, \texttt{EVENT\_STUDY}),
\texttt{family\_subtype}, \texttt{core} and a family-specific
\texttt{family\_module}. The \texttt{core} block has 13 required fields
(\texttt{estimand}, \texttt{causal\_exposure\_concept}, \texttt{outcome},
\texttt{identifying\_variation}, \texttt{sample\_definition},
\texttt{unit\_of\_analysis}, \texttt{time\_structure},
\texttt{validity\_conditioning}, \texttt{fixed\_effects}, \texttt{controls},
\texttt{clustering}, \texttt{target\_coefficient},
\texttt{identification\_assumptions}) and four optional ones
(\texttt{comparison\_group}, \texttt{variable\_transformations},
\texttt{weights}, \texttt{auxiliary\_reduced\_form\_estimand}). Every freeze
version hashes schema v1.0; the v1.1--v1.2 revisions (nullable fields; an
\texttt{OTHER} event-study subtype) postdate the panel evidence freeze and
leave this inventory otherwise unchanged.
\end{sloppypar}

\begin{sloppypar}
\paragraph{Critical fields for $V_2$.}\label{sec:appendix-protocol-fields}
The design-layer instrument profiles 16 fields, each \textsc{match},
\textsc{material mismatch}, \textsc{not applicable} or \textsc{unresolved},
with no aggregate score. Seven are critical in the pilot, where any critical
mismatch fails the run and a critical \textsc{unresolved} caused by benchmark
ambiguity excludes it (CDR path read in brackets where not eponymous):
\texttt{family} [mechanical], \texttt{estimand},
\texttt{target\_coefficient}, \texttt{treatment\_identity}
[\texttt{family\_module}], \texttt{transformations\_and\_scale}
[\texttt{core.variable\_transformations}], \texttt{comparison\_assignment}
[\texttt{core.comparison\_group}, \texttt{core.sample\_definition}] and
\texttt{inference\_convention} [\texttt{core.clustering}]; the other nine are
profile-only. In the panel the frozen scoring rule names five scored critical
fields, all adjudicated: \texttt{inference\_convention} left the pass/fail
set (rule R1 below) but stays in the critical field-verdict ledger, which
thus spans six adjudicated fields per commitment. The mechanical
\texttt{family} comparison remains scored but sits outside that ledger:
label-only differences over a form-identical estimating equation are
overridden to \textsc{match}, and every remaining family mismatch coincides
with a commitment already carrying countable recast mismatches.
\end{sloppypar}

\paragraph{Stage-1 commitment contract.}\label{sec:appendix-protocol-stage1}
Every provider call carries one system prompt (``You are an applied
econometrician. You implement research designs exactly as specified and
report what your implementation produces.'') and one user message. At Stage~1
that message holds an instruction block (``Commit to a causal design. Say
what you would estimate and how you would identify it, as a structured
representation, BEFORE seeing any data.''), three context fields (research
question, data description, institutional context) and an answer contract.
The context is a rendered, hashed per-task artifact written under two stop
rules: it never asserts that identification assumptions hold and never
reports a realised first stage, result direction or significance. A lint
finds no method name in any of the 23 contexts, and task identifiers are
opaque.

The answer contract asks for one JSON object in the schema's shape; the
verbatim templates (the Stage-1 contract alone is 35 lines) are hashed with
the prompt builder and are in the released archive. The validity gate is
deliberately narrow (\texttt{family} is one of the four, \texttt{core} is
an object, and \texttt{core.estimand}, \texttt{core.outcome} and
\texttt{core.target\_coefficient} are non-empty) so that any other
omission is measured rather than rejected. A valid commitment is hashed at
once (SHA-256 over canonical JSON) inside a function that resolves no seed
path, data dictionary or data schema. There is no retry or repair: a
gate-failing response is a recorded Stage-1 failure whose main arm does not
run, for which the gold CDR is never substituted, and which is \emph{not
scorable} at $V_2$, never a design failure.

\paragraph{Stage-2 execution contract.}\label{sec:appendix-protocol-stage2}
Stage~2 is a second, stateless provider call, not a continued chat session:
continuity is carried by the commitment, re-presented verbatim under the
heading ``THE DESIGN YOU COMMITTED TO'' and recorded by hash on the main-arm
result. The message adds the twin's data dictionary, a data schema (column
names, dtypes and row count: the shape of the data, never its values) and
the output contract. The model returns one self-contained Python script, run
as \texttt{python your\_script.py --data <one-seed.parquet>} separately on
each of the 20 seed files, which must exit~0 after printing one JSON object
with \texttt{point\_estimate} and \texttt{standard\_error}. It may import
numpy, pandas and pyarrow and nothing else beyond the standard library, must
read only the file passed, must not require network access, and must run in
under two minutes per seed. Every call has an 8{,}192-token output cap and
the mode is single-shot: a script that fails to run is a recorded failure,
not a retry. Scripts execute in a network-isolated, read-only container
holding only an interpreter and those three libraries, and before any live
call the runner executes an adversarial escape probe (sealed truth, host
credentials, network) and refuses to proceed unless every vector is closed.

\paragraph{Oracle arm.}\label{sec:appendix-protocol-oracle}
The third call never invokes Stage~1 and runs whether or not the model's own
commitment was valid. Under the heading ``DESIGN SPECIFICATION'' it
substitutes the \emph{model-facing view} of the gold CDR (the gold design
payload, unchanged, plus an \texttt{execution\_conventions} block) for the
committed design; every other prompt section, the system prompt, token cap,
model configuration, twin, seeds, sandbox and scorer are held fixed. One
asymmetry belongs to the contrast: that block pins the standard-error form,
degrees of freedom, sample rules and units, whereas a committed CDR carries
conventions only insofar as the model wrote them.

\paragraph{Scoring: $V_4$ bands.}\label{sec:appendix-protocol-scoring}
The bands are frozen in the planting registry and read, as written there,
\texttt{point\_band}: $|\mathrm{est} - \mathrm{canonical}| < 0.5 \times
SE_{\mathrm{canonical}}$ and \texttt{se\_ratio\_band}: $SE_{\mathrm{model}} /
SE_{\mathrm{canonical}} \in [0.8, 1.25]$; a seed is recovered only if both
hold on that seed's canonical values. The point band sits strictly below the
$2 \times SE$ displacement required of every registered wrong-specification
probe; the 50\% relative tolerance of the v1 release is not reused. Two
registry-named variants exist: a diagnostic twin is scored on its
standardized statistic ($|t_{\mathrm{model}} - t_{\mathrm{canonical}}| <
0.5$) plus 5\%-level test agreement, and a two-endogenous twin on one
coefficient only. A recitation diagnostic flags an estimate that misses the
point band yet lies within $0.5 \times SE_{\mathrm{canonical}}$ of a
comparable published value.

\paragraph{Scoring: $V_3$ and denominators.}
A seed is \emph{executed} when the script exits~0 and prints a JSON object
with numeric \texttt{point\_estimate} and \texttt{standard\_error}. Four
denominators stay separate: \emph{attempts} (model $\times$ pair; 207);
\emph{scored commitments} for $V_2$ (194 schema-valid less the seven of the
reference-defective pair; 187); \emph{eligible seeds} for $V_3$ ($20 \times$
the Stage-1-valid commitments; an arm that never ran counts as 20 failed
seeds); and \emph{executed seeds} for $V_4$, undefined rather than zero when
nothing executed. These units were fixed before the first panel verdict;
seeds are never independent observations.

\paragraph{$V_2$ adjudication: pilot instrument.}\label{sec:appendix-protocol-adjudication}
Free-text fields take their verdict from an adjudication log recorded as
data (agent-drafted rulings reviewed by an author), never from a model judging
at run time; each of the 90 pilot rulings is
hash-bound to the values it was made on, and each run is compared against the
gold in force when it executed (freeze v1.3 for one run, v1.4 for five).

\paragraph{$V_2$ adjudication: outcome-blinded panel batches.}
The panel contract was frozen before the first verdict and leaves the
instrument unchanged. The adjudicator reads batch files containing only the
gold design payload and anonymised commitments (no model identity, no
$V_3$/$V_4$, no oracle outcome, no usage), with the label-to-model mapping
in a sealed file. There are 23 batches, one per pair, each with up to nine
commitments in seeded-shuffled order, so that one gold's field semantics are
ruled consistently across models. Uncovered questions are resolved together
in a dated addendum of general rules, with equivalence argued on
construction, never on reasonableness. Unblinding comes last, after the
reliability audit and the $V_2$ profile freeze; later corrections must appear
as labelled post-unblind errata.

\paragraph{$V_2$ adjudication: reference review.}
A field on which at least eight of nine commitments in a batch receive the
same mismatch is automatically registered for review of the \emph{gold}:
an instrument-anomaly signal, never a verdict change; 30 flags were reviewed.
Each source reviewer receives only the pair identifier, flagged field, gold
claim, frozen Stage-1 context and archived source (never the commitments
or mismatch counts) and returns quotations, not verdicts. Source fidelity
(verified, defect, ambiguous) and model-facing determinability (determinate,
underdetermined) are ruled in separate columns: a verified but
underdetermined gold makes the field \emph{not scorable}; a confirmed defect
excludes the pair, layer-scoped, with no post-hoc gold repair. Verdict files
are never rewritten.

\paragraph{$V_2$ adjudication: benchmark-underdetermined removals.}
The eight-of-nine threshold was a review trigger, not a scoring criterion:
once review found mismatch classes arising from information absent from the
model-facing context, the two-column rule was applied to every mismatch of
the same semantic class however many models exhibited it (``disposition
follows information availability, not flag frequency'') under three rules
frozen before model identities or downstream outcomes were accessed. R1:
after verifying that none of the 23 frozen contexts pins an inference
convention, that field exits $V_2$ pass/fail globally ($V_4$ still scores SE
recovery). R2: each unflagged mismatch class receives its flagged
counterpart's ruling item by item, with no blanket void. R3: every surviving
critical mismatch is re-examined individually and is countable only on stated
grounds (family recast; misread of a stated estimand; violated
context-visible sample rule; internal incoherence; rewritten visible
identification structure; ignored stated transformation). The ledger closes
exactly: of $194 \times 6 = 1{,}164$ critical field-verdicts, 760 are
matches, 75 countable mismatches, 318 voids (160 reference review, 46 R1, 112
R2/R3), 8 unresolved and 3 not applicable.

\paragraph{$V_2$ adjudication: re-adjudication audit.}
The 614 critical-field \textsc{match} verdicts that remain scored after the
overlays form the eligible population; a 20\% random sample of 123 was
re-adjudicated, with the seed (20260823) and sampling rule fixed at the
two-batch checkpoint and the draw made only after every verdict was locked.
Three of the 123 disagreed (97.6\% agreement, 120 of 123), and all
three were corrected to countable mismatches, included in the 75 above. A
separately reported stress audit re-reviewed all 17 hedge or deferral matches
and found no disagreement.

\paragraph{Amendment record.}\label{sec:appendix-protocol-amendments}
Superseded manifests are never rewritten. The initial freeze (v1.0) has six
dated amendments. Four are dated 2026-08-18, the first three before any
paired evaluation call: v1.1 removed a metadata block carrying a legacy
identifier from all 23 model-facing views; v1.2 rewrote the institutional
context of 21 of 23 tasks, found to state identification arguments or source
results as premises, and made Stage-1 inputs hashed per-task artifacts; v1.3
implemented the three-call orchestration; v1.4, triggered by the first paid
pilot run, changes only what is recorded, not what is measured. Two
amendments declare themselves \emph{not result-blind}: v1.5 (2026-08-20) adds
a coefficient-scale pin to one task and a units pin to another after a paid
pilot trajectory (excluded, never re-scored) exposed a source-to-twin unit
ambiguity; v1.6 (2026-08-21) replaces one task's research question and data
description after the pilot design-profile review found them describing a
companion study. Neither alters a DGP, planted effect, canonical estimator,
scoring band or archived trajectory. Separately, the design-profile
instrument lies outside the twin-set freeze and is itself not result-blind
for the pilot; its panel extension is outcome-blinded by construction.

\section{Paired Study: Additional Details}\label{sec:appendix-panel-details}

The protocol itself is drawn in \cref{fig:v2-overview} (panels C and D); this
appendix records the adjudication, model settings and pilot material behind
\cref{sec:panel}.

\paragraph{Adjudication of the design layer.}
The full panel runs all 23 pairs across nine models under twin freeze v1.6:
207 Stage-1 attempts, of which 13 produce no schema-valid commitment and are
recorded as not scorable (never as design failures), leaving 194 committed
designs. Design-specification verdicts were adjudicated outcome-blinded in
per-pair batches with model identities sealed. A reference review by
commitment-blinded reviewers (language-model agents, \cref{sec:threats})
then verified every flagged gold claim against the archived source and the
frozen Stage-1 context, ruling source fidelity and model-facing
determinability in separate columns. For one pair (cvp\_023) the archived
document is a later replication study of the cited paper rather than the
paper itself; its Stage-1 context was reconstructed from that document while
its gold cites the original, and its unit-of-analysis field was recorded as a
context-fidelity defect and removed from scoring. Mismatches that concerned conventions absent
from the model-facing context (estimator-column anointments, footnote
clustering keys, instrument lag structures) were removed from scoring as
benchmark-underdetermined rather than charged as model error, uniformly and
irrespective of how many models exhibited them; the inference-convention
field left the scored set entirely, since no frozen context states one, and
is reported as a descriptive alignment profile. One pair shipped a defective
gold (a literal draft marker that also reached its oracle prompts) and is
excluded layer-scoped. A seeded 20\% re-adjudication of eligible \textsc{match}
verdicts agreed at 97.6\% (120 of 123 re-adjudicated, from 614
eligible; three corrections, all adopted); a
100\% re-review of hedge/deferral verdicts agreed at 17/17.

\begin{sloppypar}
\paragraph{Models and inference settings.} The primary panel evaluates
nine models under identical contracts (endpoint identifiers as archived in
the per-call ledgers): \texttt{claude-opus-4-6}, \texttt{claude-sonnet-4-6},
\texttt{gpt-5}, \texttt{gpt-4o}, \texttt{o3}, \texttt{gemini-2.5-flash},
\texttt{moonshot-v1-128k}, \texttt{Qwen/\allowbreak Qwen3-235B-\allowbreak A22B-\allowbreak Instruct-2507}, and
\texttt{deepseek-ai/\allowbreak DeepSeek-V4-Pro}, all run in August 2026 with
provider-default sampling settings, single-shot with no retry or repair,
three calls per pair (Stage-1 commitment, main arm, oracle arm), an
8{,}192-token output cap, and an execution contract restricted to
numpy/pandas/pyarrow and the Python standard library. Endpoint
identifiers, per-call token ledgers, prompt hashes and submitted scripts are in
the released trajectory archive; the identifiers are provider aliases and the
ledgers carry no timestamps.
Of the 35 paired calls that stopped at the 8{,}192-token output cap, 24 were Gemini, 9 GPT-5 and 2 o3; the latter two spent the whole budget on hidden reasoning tokens.
\end{sloppypar}

\begin{figure}[h]
\centering
\includegraphics[width=0.78\textwidth]{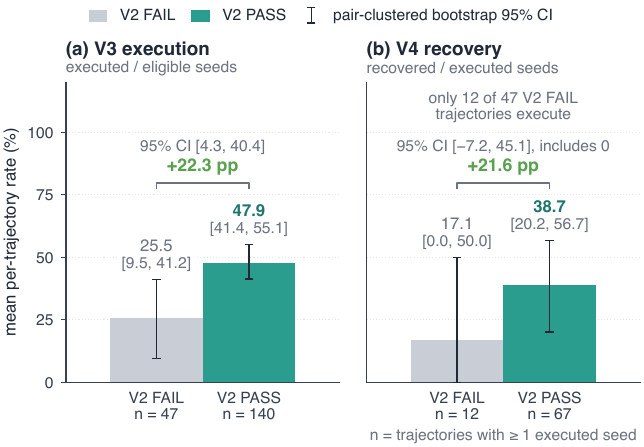}
\caption{Conditional downstream execution ($V_3$) and recovery ($V_4$) by
V2 status, with pair-clustered bootstrap 95\% CIs~\citep{efron1994bootstrap}. The recovery difference's
interval includes zero (12 executing V2-fail trajectories); the association
is directionally consistent, not resolved. $V_2$ status is not randomized;
these are conditional associations, not a causal decomposition.}
\label{fig:panel-conditional}
\end{figure}

\subsection{A two-arm pilot on six pairs}\label{sec:pilot}

Before the full panel, six of the 23 pairs were run end to end with one model
(\texttt{gpt-5.6-sol}, the archived ledger identifier for the pilot-era GPT-5.6 endpoint; single-shot, both arms; three provider calls per pair).
Every trajectory is archived under a hashed evidence freeze; after the
telemetry contract was expanded mid-programme, runs additionally preserve the
committed CDR, call-level usage, the submitted scripts and per-seed numerics.
The earlier Bernanke--Kuttner run predates that expansion and retains only the
categorical verdicts of note~$a$. \Cref{tab:pilot} reports the
result as a layer profile per pair, not a success rate: with six trajectories
the deliverable is where each chain held or broke.

\begin{table}[h]
\centering
\caption{Layer profile of the six-pair pilot. $V_1$: committed family correct.
$V_2$: critical-field mismatches in the committed CDR against the
contemporaneous gold CDR (seven critical fields adjudicated
\textsc{match}/\textsc{mismatch} per field, non-critical profile mismatches
in the notes, no aggregate score). $V_3$: seeds
executed (main arm). $V_4$: executed seeds whose estimate and SE fall inside
the frozen bands. The oracle arm reruns Stage~2 with the gold CDR.}
\label{tab:pilot}
\small
\begin{tabular}{@{}lcllll@{}}
\toprule
Pair & $V_1$ & $V_2$ profile & $V_3$ main & $V_4$ main & Oracle $V_3\,/\,V_4$ \\
\midrule
Bernanke--Kuttner & \checkmark & 0 critical & 20/20 & 20/20$^{a}$ & 20/20 \\
AJR & \checkmark & 0 critical$^{b}$ & 19/20$^{c}$ & 19/19 & 20/20 \\
ADH & \checkmark & excluded$^{d}$ & 20/20 & n/a$^{d}$ & 20/20 \\
Oregon HIE & \checkmark & 0 critical$^{b}$ & 20/20 & 20/20 & 20/20 \\
Dell (Mita) & \checkmark & \textbf{4 critical} & 20/20 & \textbf{0/20} & 20/20 \\
KLM & \checkmark & \textbf{5 critical} & \textbf{0/20}$^{e}$ & n/a & 20/20 \\
\bottomrule
\end{tabular}

\raggedright\footnotesize
$^{a}$Boolean verdicts preserved; this run predates the expanded telemetry
contract, so its per-seed numerics are recorded as permanently unknown rather
than backfilled.
$^{b}$Non-critical profile mismatches recorded (AJR~3, Oregon~2); see text.
$^{c}$One seed lost to a container abort: execution-conditioned success is
19/19 and end-to-end seed completion 19/20; the missing seed is
infrastructure, not model behaviour.
$^{d}$Benchmark-side unit-contract ambiguity discovered by this run and
resolved prospectively by a dated amendment; the trajectory is excluded from
model-performance inference in both directions.
$^{e}$The model's own feasibility guard: its committed market-model estimation
window ($-250$ to $-31$, $\geq$120 observations) cannot exist in a twin
shipping 60 trading days per firm, and the script raised rather than
substituting an uncommitted design.
\end{table}

Three regularities in the profile are worth stating plainly, without
generalizing beyond $n{=}6$. First, the \emph{oracle arm is uniform}: with
the gold CDR substituted, every seed of every pair executes and scores inside
the frozen bands. Under this protocol that is a substitution contrast: the
gold-CDR substitution changed downstream execution and recovery wherever the
arms diverge. It is not a strict causal decomposition of error sources. Second, \emph{family recognition is not the bottleneck}: all six
committed CDRs name the correct design family; every failure occurs downstream
of $V_1$. Third, the four distinct main-arm outcomes form a small taxonomy of
\emph{observed trajectory types}, two of which are not failures at all,
rather than a rate of how often anything succeeds:

\begin{itemize}
  \item \textbf{Full chain} (Bernanke--Kuttner, Oregon HIE): no
        critical-field mismatch in the committed CDR, execution on every seed,
        and recovery inside the frozen bands on every executed seed
        ($V_2 \rightarrow V_3 \rightarrow V_4$). For Bernanke--Kuttner the
        recovery claim is categorical (note~$a$): the preserved verdicts, not
        per-seed values.
  \item \textbf{Design divergence, faithfully propagated} (Dell): the
        committed CDR differs from the gold on four critical fields at once
        (target specification, running variable, band and analysis unit,
        inference), adjudicated from the committed fields alone, before any
        reference to outcomes; the archived script implements that commitment
        on every seed and is wrong on every seed: sign-unstable across seeds,
        standard errors 16--28$\times$ the canonical. Surface-plausibility and
        runnability checks would not establish correctness here;
        reference-based numerical validation exposes the discrepancy.
  \item \textbf{Infeasible commitment, model-generated guard halt} (KLM):
        the commitment reproduces a recognizable source-paper event-study
        specification, including a pre-event estimation window, but diverges
        from the benchmark gold on five critical fields and names an
        estimation window the twin's data geometry cannot contain; Stage~1
        closes before the data dictionary is visible, and the single-shot
        contract offers no renegotiation. A model-generated feasibility guard
        then caused a submission-level execution failure on every seed. The
        taxonomy records the mechanism (a guard, not crashing code) as a
        distinction of kind, not a credit.
  \item \textbf{Contract ambiguity, charged to the benchmark} (ADH): the
        model implemented exactly the per-\$1{,}000 estimand it committed,
        against a twin whose canonical execution treats the shipped exposure
        as an already-normalized coefficient unit, a correspondence the
        contract in force did not explicitly bridge; estimates and
        SEs scale together by $\sim$1{,}055$\times$ and the $t$-statistic is
        nearly unchanged, so every text-level check passes. The run is
        excluded from model-performance inference in both directions (it is
        neither a model failure nor, under the later clarified contract,
        retroactively a pass) and its lesson kept: the mismatch is
        invisible to any check short of comparing coefficient magnitude to an
        executable reference.
\end{itemize}

AJR sits between the taxonomy's poles and illustrates why $V_2$ is a profile
rather than a score: the commitment adds conditioning (latitude, continent
effects) that the gold's anointed specification excludes (three
non-critical field mismatches and no critical-field mismatch), yet the twin
ships no such columns,
the implementation estimates the realizable core design, and recovery passes
on every executed seed. A scalar $V_2$ score would have averaged this into
noise; the profile keeps both facts.

The $V_2$ instrument and its 90-entry adjudication log were built after these
outcomes were known and are recorded as such: not result-blind, with
structural mitigations (field-semantics verdicts, rationales that stand
without reference to $V_4$, no code path from the profiler to execution
output, and per-run binding to the contract version in force at execution
time).

\subsection{The four-task prototype across nine models}\label{sec:prototype}

The pilot's protocol descends from a smaller prototype run earlier in the project:
four tasks (one per family) under the same two-stage contract (brief, then
a committed interpretation, then data, then code and execution) with
Stage-1 scores frozen before Stage~2 and canonical results precomputed. Nine
models completed all four tasks. The prototype is targeted validation at
descriptive scale, and its value is the structure of its disagreements rather
than its rates.

\paragraph{Results.} Method recognition was near-ceiling (35 of 36 cells) and
execution high (33 of 36), yet recovery separated: the event-study and DID
tasks resolved cleanly (9/9 and 8/8 of executing models correct), while the
RDD task executed for all nine models and recovered for six. The three
misses are the instructive part: two vendors' models (GPT-4o and DeepSeek)
produced the \emph{same} wrong estimate, 5.075 against a canonical 1.428
($3.55\times$, relative error 2.55), and a third missed at 0.59.

\paragraph{Case study: the $3.55\times$ estimate.} Both models committed the
same specification error, an \texttt{rdrobust} call~\citep{calonico2014robust} on truncated support
that leaves roughly 21 effective observations, and both returned an
estimate that reads as an ordinary, confident RDD result. The workflow passes
a text check and a does-it-run check; only the comparison against the
canonical estimate exposes it, and the cross-vendor coincidence of the exact
value is what marks it as a systematic specification failure rather than a
sampling accident. This is the observation the paired protocol was then built
to instrument at scale.

\section{Open-Weight Runs}\label{sec:appendix-openweight}

This appendix documents the open-weight rows quoted in \cref{sec:execution}:
the protocol under which Qwen3-235B-A22B-Instruct-2507 and DeepSeek-V4
were run, one correction to the execution environment, and the full counts.
Llama-3.3-70B-Instruct is listed alongside them as a robustness check only;
it is never part of the model ranking, and the three rows are not a
benchmark of open-weight models.

\paragraph{Protocol.}
The two additional families were run after the v1 release was frozen, under
a protocol that was prespecified and frozen before the first full-panel
generation call and before any scoring. The arms are
\model{Qwen/Qwen3-235B-A22B-Instruct-2507} and
\model{deepseek-ai/DeepSeek-V4-Pro} (reported as DeepSeek-V4), both served
through a hosted inference endpoint whose resolved model string was checked
against the requested one by a single liveness call per arm. Everything else
is inherited unchanged from the v1 executable arm: the same 100 scenarios
and realised datasets, the same system and user prompts, a 4{,}096-token
output budget, provider-default sampling with no reasoning-mode parameters,
a single shot per scenario, no retry on a substantive failure, and no
per-model prompt adjustment. Scoring reuses the v1 chain without
modification: R execution, coefficient extraction, comparison with the
canonical estimate under \cref{eq:tolerance} at $\tau = 0.50$, and the
event-study rule that accepts a cumulative window of up to 11 post-event
days. Canonical estimates were read from the frozen v1 artifacts and never
recomputed. The protocol fixed the endpoints (\lexec, \lcorrect, and
$P(\text{wrong}\mid\text{executable})$ per arm, with a by-family breakdown)
and the wording of three possible readings in advance, and committed to
reporting every arm whatever it showed. Coefficient extraction for the new
outputs required new calls to the extraction model, because these outputs
are absent from the v1 extraction cache.

\paragraph{Realised sample.}
All 200 generations completed (100 per arm) with no transient retries, no
failures, and no truncated responses; every response contained an R code
block, so each arm's denominator is 100 and every non-executing cell is
code that was extracted and then failed at run time. The runs wrote only to
their own directory, and the hashes of the two v1 score tables, recorded
before the run, matched after each scoring pass.

\paragraph{Environment correction.}
The numbers in this paper come from the second of two scoring passes; the
first is superseded. To run unvetted model-written R code safely, the two
new arms were executed in a network-isolated container (R~4.4.2, no
credentials mounted) rather than on the host R installation that scored the
v1 panel. The first container was built from the \emph{documented} v1
package list, which turned out to be narrower than the package set actually
present when the v1 panel was scored. In that first pass Qwen3 executed on
17 scenarios (16 correct; 1 of 17 wrong) and DeepSeek-V4 on 46 (42 correct;
4 of 46 wrong). An audit of every non-executing cell (83 and 54) found no
code-extraction failures; 73 and 27 of them were ``there is no package
called'' errors, 65 of those 100 for \texttt{readr} alone and the rest for
other routine packages that, by the protocol record, v1-panel workflows had
loaded successfully. This is a harness deviation, not model behaviour, and
it made the first-pass counts non-comparable with the v1 panel. The remedy
was written into the protocol record before re-scoring: a derived image
adds exactly the eleven packages evidenced in the v1 environment
(\texttt{readr}, \texttt{tidyverse}, \texttt{rddensity},
\texttt{lubridate}, \texttt{ivreg}, \texttt{plm}, \texttt{estimatr},
\texttt{quantreg}, \texttt{modelsummary}, \texttt{foreign},
\texttt{knitr}). \texttt{rdd} and \texttt{stargazer} were deliberately
\emph{not} added: they were absent from the v1 environment too (the v1
score table records 15 and 24 failures on them), so failing on them is a
genuine failure under v1 semantics. Both arms were re-scored uniformly with
the unchanged chain, no output was regenerated, and the first-pass scores,
summary, and execution logs are archived. The correction is post hoc and
not result-blind: it was triggered by the first-pass results, and it raises
\lexec\ from 17 to 35 (Qwen3) and from 46 to 59 (DeepSeek-V4). It barely
moves the conditional quantity the main text relies on (1/17 to 1/35; 4/46
to 5/59), and no cell that executed in the first pass changes status: the
18 and 13 newly executing cells add 18 and 12 correct workflows. The 26
missing-package failures that remain (25 Qwen3, 1 DeepSeek-V4) all name
packages recorded as absent from the v1 environment as well (\texttt{lfe}~11,
\texttt{ivmodel}~4, \texttt{stargazer}~4, \texttt{rdd}~4,
\texttt{rddtools}~2, \texttt{ivpack}~1), so one iteration closed the gap.
The v1 documentation's package list therefore understates the environment
in which the v1 panel was actually scored.

\paragraph{Results.}
\Cref{tab:appendix-openweight-main} gives the counts behind the main-text
rates. The Llama row is not a new run: it is read directly from the frozen
v1 score table, where Llama was scored in the v1 environment, and it was
not re-executed in the container. The
conditional rates rest on small executing denominators (35, 59, and 41
workflows), and the intervals in the last column are correspondingly wide;
they support the main text's qualitative contrast (Qwen3 and DeepSeek-V4
lose most workflows before execution, Llama loses about half of its
executing workflows after it) and no finer comparison.

\begin{table}[h]
\centering
\caption{Open-weight runs on the 100-scenario executable arm. ``Scen.'' is
the number of scenarios; ``Wrong'' is the executed-but-wrong count (\lexec\
passed, \lcorrect\ failed); $P(\text{wrong}\mid\text{exec.})$ divides it by
the \lexec\ count. Brackets are Wilson 95\% intervals. Qwen3 and DeepSeek-V4
were run after the v1 release was frozen (corrected environment).
$^{\dagger}$Robustness only: the Llama row is read from the frozen v1 score
table and is never part of the model ranking. Among the two ranked runs, the
better value in each outcome column is in \tA{blue} and the other in
\tB{purple} (higher is better for \lexec\ and \lcorrect, lower for the two
error columns); the robustness row is not coloured.}
\label{tab:appendix-openweight-main}
\small
\setlength{\tabcolsep}{5pt}
\begin{tabular}{@{}lccccl@{}}
\toprule
\textbf{Model} & \textbf{Scen.} & \lexec & \lcorrect &
\textbf{Wrong} & $P(\text{wrong}\mid\text{exec.})$ \\
\midrule
\grouprow{6}{Post-freeze runs (corrected environment)}
Qwen3-235B-A22B-Instruct-2507 & 100 & \tB{35} & \tB{34} & \tA{1} & \tA{2.9\%} [0.5, 14.5] \\
DeepSeek-V4 & 100 & \tA{59} & \tA{54} & \tB{5} & \tB{8.5\%} [3.7, 18.4] \\
\midrule
\grouprow{6}{Robustness only (frozen v1 score table)}
Llama-3.3-70B-Instruct$^{\dagger}$ & 100 & 41 & 20 & 21 &
51.2\% [36.5, 65.7] \\
\bottomrule
\end{tabular}
\end{table}

\paragraph{By design family.}
\Cref{tab:appendix-openweight-family} splits the same counts by design
family (30 DID, 24 event-study, 24 IV, and 22 RDD scenarios). Cells are
small and are reported for completeness, as the protocol required, not as a
family-level comparison. Two patterns are visible without over-reading
them: Qwen3's execution failures are concentrated in RDD (1 of 22
executes), and for both new arms the executed-but-wrong cases fall mainly in
the event-study family (1 of 1 for Qwen3, 3 of 5 for DeepSeek-V4), whereas
Llama's are spread across all four families.

\begin{table}[h]
\centering
\caption{Open-weight runs by design family. Each cell is
\lexec\,/\,\lcorrect\ out of the $n$ scenarios in that family; the
difference between the two is the executed-but-wrong count. Highest
\lcorrect\ count per column in \tA{blue}; the robustness-only Llama row is
excluded from the comparison.}
\label{tab:appendix-openweight-family}
\small
\begin{tabular}{lcccc}
\toprule
\textbf{Model} & \textbf{DID} ($n{=}30$) & \textbf{ES} ($n{=}24$) &
\textbf{IV} ($n{=}24$) & \textbf{RDD} ($n{=}22$) \\
\midrule
Qwen3 & 10\,/\,10 & 16\,/\,\tA{15} & 8\,/\,8 & 1\,/\,1 \\
DeepSeek-V4 & 16\,/\,\tA{16} & 16\,/\,13 & 18\,/\,\tA{17} & 9\,/\,\tA{8} \\
\midrule
Llama (robustness only) & 10\,/\,3 & 16\,/\,9 & 8\,/\,4 & 7\,/\,4 \\
\bottomrule
\end{tabular}
\end{table}

\section{Repair Pilot}\label{sec:appendix-repair}

This appendix gives the design and full counts of the one-round repair pilot
summarized in \cref{sec:boundary}. It is a pilot (three models, one
repair round, one fixed feedback sentence per failure type, several thin
cells) run after the v1 release was frozen; it changes no v1 number.

\paragraph{Sampling frame.}
The frame is every workflow in the frozen v1 score table that fails
\lcorrect\ for three models fixed in advance as a high, a middle, and a low
ability band of the v1 panel (GPT-5, o3, and Kimi): 172 failed
workflows in all (28, 54, and 90). It is stratified by design family (DID, event study, IV, RDD) and
failure type: \emph{non-executing} (\lexec\ failed) or
\emph{executed-but-wrong} (\lexec\ passed, \lcorrect\ failed). Within each
model $\times$ family $\times$ type cell all failures are taken up to a cap
of 10; the five larger cells are subsampled with a fixed seed. The realised
sample, frozen before any repair call, is 136 workflows: all 28 GPT-5
failures, 42 of o3's, and 66 of Kimi's; 100 non-executing and 36
executed-but-wrong. Every sampled workflow contains an extracted R code
block, so ``non-executing'' uniformly means code that was present and failed
at run time. Because of the cap, the rates below describe this stratified
sample, not the population of all 172 failures.

\paragraph{The single repair round.}
Each case is one conversation: the original v1 prompt for that scenario,
byte-identical; the model's original cached response, byte-identical; and
then exactly one fixed sentence chosen by failure type. For a non-executing
workflow the model is told, verbatim, ``The submitted workflow failed
execution. You may revise it once.'' For an executed-but-wrong workflow it
is told ``The workflow executed, but the extracted causal estimate did not
satisfy the benchmark's numerical correctness criterion. You may revise it
once.'' Nothing else is added: no error log, no canonical estimate, no
tolerance, no sign or size of the error, no reference code, and no
description of the scoring mechanism. Each case receives one repair
response; only transport-level errors are retried (at most twice, with the
identical request), and a case that still returns nothing is recorded as no
response and counted as unrepaired.

\paragraph{Scoring rule.}
A workflow counts as \emph{repaired} if and only if its revised response
passes \lcorrect\ under the unmodified v1 scoring chain: the revised R code
is extracted and executed (60-second limit), the estimate is extracted by
the same extraction model, and it must fall within the default band of
\cref{eq:tolerance} ($\tau = 0.50$) around the canonical estimate, with the
v1 event-study window rule applied to event-study cells. The primary
endpoint is the share of originally failing workflows that are repaired,
with Wilson 95\% intervals. The protocol fixed four secondary cuts (model,
failure type, design family, mechanical versus conceptual repair) and the
reading of either a high or a low repair rate before any repair call~\citep{olken2015promises}.

\begin{table}[h]
\centering
\caption{Repair pilot: share of originally failing v1 workflows that pass
\lcorrect\ after one repair round, pooled and by the prespecified cuts.
Brackets are Wilson 95\% intervals. No-response cases count as unrepaired.
Top three rates among the cut rows in \tA{blue}, \tB{purple}, \tC{orange};
the bold pooled row is the headline and is not ranked.}
\label{tab:appendix-repair-main}
\small
\setlength{\tabcolsep}{5pt}
\begin{tabular}{llccc}
\toprule
\textbf{Cut} & \textbf{Group} & $n$ & \textbf{Repaired} &
\textbf{Rate [95\% CI]} \\
\midrule
\textbf{Pooled} & \textbf{all sampled failures} & \textbf{136} & \textbf{41} & \textbf{30.1\% [23.1, 38.3]} \\
\midrule
Model & GPT-5 (high) & 28 & 18 & \tA{64.3\% [45.8, 79.3]} \\
 & o3 (middle) & 42 & 15 & \tC{35.7\% [23.0, 50.8]} \\
 & Kimi (low) & 66 & 8 & 12.1\% [6.3, 22.1] \\
\midrule
Failure type & non-executing & 100 & 33 & 33.0\% [24.6, 42.7] \\
 & executed-but-wrong & 36 & 8 & 22.2\% [11.7, 38.1] \\
\midrule
Design family & DID & 38 & 10 & 26.3\% [15.0, 42.0] \\
 & event study & 33 & 13 & \tB{39.4\% [24.7, 56.3]} \\
 & IV & 28 & 7 & 25.0\% [12.7, 43.4] \\
 & RDD & 37 & 11 & 29.7\% [17.5, 45.8] \\
\bottomrule
\end{tabular}
\end{table}

\paragraph{Model by failure type, and incidents.}
\Cref{tab:appendix-repair-cross} crosses the two cuts the main text reads
together. The pooled asymmetry (22.2\% against 33.0\%) is not uniform across
models: GPT-5 and o3 repair non-executing workflows more often than
executed-but-wrong ones (o3 repairs none of its four), while Kimi shows the
reverse on a larger cell. GPT-5 and o3 each have only four
executed-but-wrong failures in the whole v1 panel, and the two pooled
intervals in \cref{tab:appendix-repair-main} overlap, so the asymmetry is a
pilot-level observation. Three execution facts are disclosed. (i)~132 of 136
repair calls returned a response, each containing R code; the other four
(all o3, all non-executing: one RDD and three IV scenarios) failed at the
transport level after both permitted retries and count as unrepaired, as
the protocol required. (ii)~The availability check that precedes the repair
calls first misclassified o3 as unavailable, because its small output
budget was consumed by the model's internal reasoning; the budget was
raised, all three models passed, and the o3 cases were run later under the
unchanged protocol. The protocol's literal rule cancels an arm that fails
this check, so this is a recorded deviation. (iii)~Unlike the open-weight
protocol of \cref{sec:appendix-openweight}, the freeze of this protocol and
sample carries no independent timestamp; it is self-attested.

\begin{table}[h]
\centering
\caption{Repair pilot: repaired\,/\,$n$ by model and original failure type.
Cells with $n = 4$ are the complete set of such failures for that model in
the v1 panel. Largest fraction per column in bold.}
\label{tab:appendix-repair-cross}
\small
\setlength{\tabcolsep}{6pt}
\begin{tabular}{lccc}
\toprule
\textbf{Original failure type} & \textbf{GPT-5} & \textbf{o3} &
\textbf{Kimi} \\
\midrule
Non-executing & \textbf{16\,/\,24 (66.7\%)} & \textbf{15\,/\,38 (39.5\%)} &
2\,/\,38 (5.3\%) \\
Executed-but-wrong & 2\,/\,4 (50.0\%) & 0\,/\,4 (0.0\%) &
\textbf{6\,/\,28 (21.4\%)} \\
\bottomrule
\end{tabular}
\end{table}

\paragraph{Mechanical versus conceptual repairs.}
Each of the 41 successful repairs is classified by a fixed syntactic rule
applied to the original and the revised R code. A pattern set detects which
estimator-call families appear in each script: fixed-effects panel calls
(\texttt{feols}, \texttt{felm}, \texttt{plm}), plain \texttt{lm},
instrumental-variable calls (\texttt{ivreg}, \texttt{iv\_robust},
\texttt{tsls}), regression-discontinuity calls (\texttt{rdrobust},
\texttt{RDestimate})~\citep{calonico2014robust}, and staggered-DID package calls (\texttt{att\_gt})~\citep{callaway2021did}.
A repair is \emph{conceptual} if the set of detected families changes
between the two scripts and \emph{mechanical} if it does not, which covers
syntax, library, variable-name, and output-formatting fixes. The rule labels
32 repairs mechanical and 9 conceptual. The protocol allowed manual
adjudication of ambiguous cases; the frozen labels coincide with the pattern
rule in all 41 cases, so the split reported here is the rule's output. It is
a coarse proxy in both directions: a changed call family need not be a
changed identification strategy (in 8 of the 9 conceptual cases the only
family detected in the revised script is plain \texttt{lm}), and an
unchanged family does not rule out a changed specification. The split
indicates, as the main text uses it, that one round of knowing-you-failed
mostly fixes broken code; it does not measure how often models revise their
identification reasoning.

\section{Prospective Replication: Six New Pairs}\label{sec:appendix-v3}

\begin{figure}[t]
\centering
\includegraphics[width=0.7\textwidth]{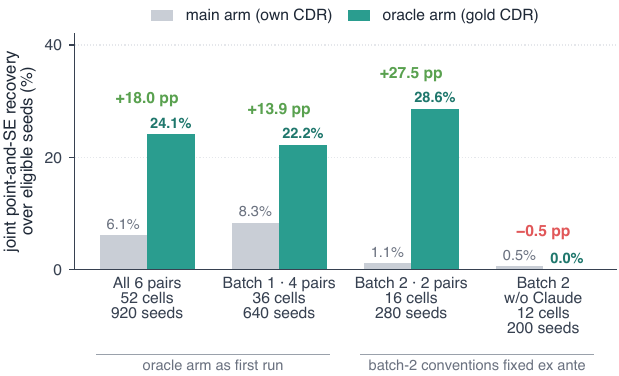}
\caption{Prospective replication on six pairs built after the claim under a
protocol frozen in advance (\cref{tab:appendix-v3}): joint point-and-SE
recovery over eligible seeds, main arm (grey) vs.\ oracle arm (teal), for all
six pairs, batch 1 (four pairs, oracle arm as first run), batch 2 (two pairs,
conventions fixed before any call) and batch 2 without the two Claude-family
models, whose 20/20 oracle seeds on both pairs are the entire batch-2 gain.}
\label{fig:v3-prospective}
\end{figure}

\paragraph{Construction and runs.} The six pairs were built after the
23-pair panel had been analysed, under a construction protocol frozen in
advance: a gold design frozen before the twin is built, a twin whose sealed
one-shot acceptance run is never redrawn, and two independent
re-implementations of the canonical estimator that must agree on all 20
seeds. Runs used the panel's contract unchanged (single-shot, no retry, up to
three calls per cell, 8{,}192-token cap, same sandbox and scorer). Batch~1
(four pairs, nine models, 36 cells) ran in August 2026. Its oracle arm showed
cross-vendor \emph{identical} off-canonical estimates on two pairs, the
signature of a convention the oracle view left free, and was re-run twice
after pins restating frozen artifacts were added (a units pin for all four
pairs, then an instrument-realisation pin and a correction-formula pin for
the two affected pairs). Those pins are post hoc, so \cref{tab:appendix-v3}, \cref{fig:v3-prospective}
reports the oracle arm as first run, after the units pin, and under the
frozen splice. Batch~2 (two pairs, 16 cells, 46 calls, September 2026) fixed
every such convention in the oracle view \emph{before} any call, from the
same frozen bindings the independent re-implementers had received; its
primary summary and reporting commitment were committed before the first
call. One panel endpoint (\model{moonshot-v1-128k}) had been retired by its
provider and returns 404; its two cells are not measured and no successor was
substituted.

\begin{table}[h]
\centering
\caption{Joint point-and-SE recovery over eligible seeds ($20\times$
schema-valid commitments) on the six new pairs. ``Without Claude'' drops the
two Claude models, whose family also built the references. Top three per
recovery column among the batch rows in \tA{blue}, \tB{purple}, \tC{orange}
(ties share a tier); the pooled headline row is bold and uncoloured.}
\label{tab:appendix-v3}
\small
\setlength{\tabcolsep}{4pt}
\begin{tabular}{@{}lcccccc@{}}
\toprule
 & & & \multicolumn{2}{c}{all models} & \multicolumn{2}{c}{without Claude} \\
\cmidrule(lr){4-5}\cmidrule(l){6-7}
 & cells & eligible & main & oracle & main & oracle \\
\midrule
Batch 1, oracle as first run & 36 & 640 & \tA{8.3\%} & 22.2\% & \tA{2.7\%} & \tB{12.5\%} \\
Batch 1, oracle after units pin & 36 & 640 & \tA{8.3\%} & \tC{25.0\%} & & \\
Batch 1, oracle frozen splice & 36 & 640 & \tA{8.3\%} & \tA{39.8\%} & \tA{2.7\%} & \tA{20.8\%} \\
Batch 2, conventions fixed ex ante & 16 & 280 & 1.1\% & \tB{28.6\%} & \tC{0.5\%} & \tC{0.0\%} \\
\midrule
\textbf{Six pairs, oracle as first run} & \textbf{52} & \textbf{920} & \textbf{6.1\%} & \textbf{24.1\%} & & \\
\bottomrule
\end{tabular}
\end{table}

\paragraph{Batch 2 by cell.} Of 16 cells, 14 produced a schema-valid
commitment (Gemini and GPT-4o each failed Stage~1 on one pair). Among valid
cells, 4 executed at least one main-arm seed and 6 at least one oracle seed.
Main arm: Opus, Sonnet and o3 each recovered 1 of 20 seeds on one pair;
nothing else recovered. Oracle arm: Opus and Sonnet recovered 20 of 20 seeds
on both pairs; o3 and DeepSeek executed on one pair and recovered none; the
remaining cells did not execute. Nine calls stopped at the token cap (GPT-5
and Gemini four each, o3 one), and Qwen3 again imported a package the
execution contract forbids.

\section{Boundary Experiments}\label{sec:boundary}

\subsection{The failure mode is not R-specific}

A matched Python-backend subset (30 scenarios; Opus, GPT-5, o3; run before
the v1 release was frozen and not included in it) reruns the same tasks with a Python execution environment. The
executed-but-wrong rate under Python is 14/58 (24.1\%), against a
matched-pair R baseline of 1/60 (1.7\%) and the primary 100-scenario DGP panel's 15.5\%; the
pairwise $2\times2$ over 88 matched pairs is 29 both-pass, 15 Python-only,
30 R-only, 14 both-fail. The subset supports exactly one claim: the
executed-but-wrong failure mode is present in both backends; it does not
establish backend invariance, and with three models the cross-backend
ordering carries no statistical meaning, so no ranking comparison is drawn
from it.

\subsection{One repair round is not enough}

A protocol-frozen repair pilot (specification fixed before scoring) takes 136 failed v1 workflows and grants one
oracle-flagged repair round: the model is told its output failed and may
resubmit once. Pooled recovery is 41/136 (30.1\%, Wilson 95\% CI
[23.1, 38.3]) and is sharply ability-graded: GPT-5 64.3\%, o3 35.7\%,
Kimi 12.1\%. Directionally, executed-but-wrong workflows are
repaired \emph{less} often than non-executing ones (22.2\% versus 33.0\%; the
intervals overlap and one model shows the reverse, \cref{sec:appendix-repair};
o3 repaired 0 of 4 executed-but-wrong cases), and among the 41 successful
repairs, 32 are mechanical fixes against 9 conceptual ones. A single round of
knowing-you-failed is largely a mechanism for fixing broken code, not for
fixing a wrong analysis. That is why the paired protocol's repair rule
stays typed and why identification-grade feedback is left to the companion
line rather than claimed here.

\subsection{Controlled repair on paired-task failures}\label{sec:appendix-repair-paired}

A second repair experiment, prespecified and committed before any call, takes
the \emph{census} of failing main-arm trajectories of the 23-pair panel (162:
108 that executed no seed, 54 that executed but did not recover on all seeds)
and grants each exactly one stateless repair call with typed feedback only:
the runtime error of the first failing seed, or the statement that the
reported point estimate and standard error fell outside the acceptance bands
on $M$ of $E$ executed seeds, with nothing about magnitude, sign or the
canonical value. Endpoints, system prompt, 8{,}192-token cap, sandbox, seeds
and scorer are those of the panel. Of the 162, 20 were not run because their
endpoint (\model{moonshot-v1-128k}) has been retired by the provider, and one
call never returned and is recorded as an infrastructure void; 141 are scored
(141 paid calls).

After repair, joint point-and-SE recovery over eligible seeds is 14.3\%
(pair-clustered 95\% CI $[8.1, 21.5]$): 16.1\% among trajectories that had not
executed and 11.1\% among those that had executed without recovering. The
prespecified directional hypothesis holds: at least one seed is recovered for
22.9\% of previously non-executing trajectories against 7.1\% of
executed-none-recovered ones ($+15.7$pp, CI $[2.6, 29.3]$). Typed feedback
repairs failures to \emph{run} more often than failures to \emph{recover}, as
in the executable arm's pilot, now with an interval that excludes zero.
Execution itself is restored for 36 of 70 non-executing trajectories but only
4 of the 20 whose original call had stopped at the token cap; 23 repair calls
stopped at the cap again. Of the 20 trajectories that went from no recovered
seed to at least one, 7 declared in the required header that the revised
script deviates from the committed design and 2 carried no parseable
declaration; this is self-report, and we make no claim about undeclared
switching. One partly recovered trajectory lost recovered seeds.

\section{Text Proxies and Calibration}\label{sec:proxies}

\begin{figure}[h]
\centering
\includegraphics[width=\textwidth]{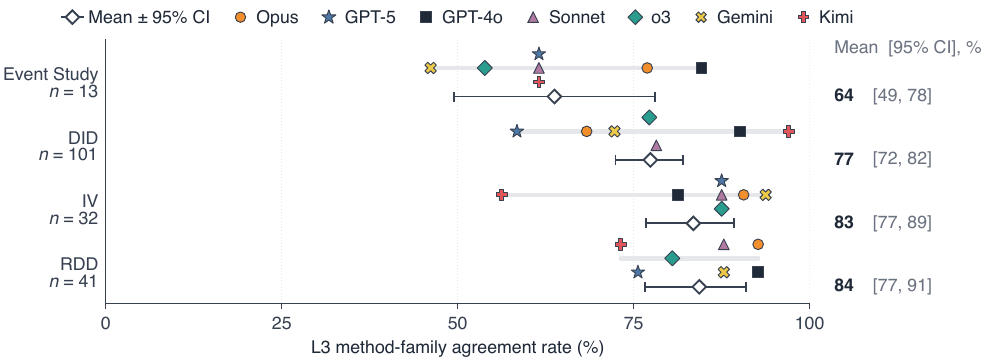}
\caption{Exp~A L3 method-family agreement by family. Markers show the seven
model-level agreement rates; diamonds and whiskers the cross-model mean with
paper-cluster bootstrap 95\% intervals. Event Study has the lowest observed
mean (64\%, $n{=}13$) and is exploratory at that corpus size. L3 is a
text-agreement diagnostic, not an executable correctness endpoint.}
\label{fig:expa_text}
\end{figure}

\paragraph{Text-agreement is label-noise-bound.} On the real-paper arm, L3
method-family agreement is scoreable for 187 of the 259 real-paper contexts
(a count unrelated to the paired panel's 187 scored commitments) and L4 direction
agreement for 92 of 259; the unscoreable remainder is dominated by papers where
the four-LLM consensus never converged on a single label. Model rankings do not
collapse to one textual ability score: GPT-4o leads L3 (88.8\%) while
Sonnet leads L4 (85.9\%), and \cref{fig:expa_text} shows the family
breakdown, with Event Study lowest (64\%, $n{=}13$; exploratory at that
corpus size). What bounds the interpretation is the blinded 30-paper human
audit, stratified by the strength of the consensus being audited: agreement
with the consensus is 100\% where at least three of the four labelers agree,
56\% under a bare plurality, and undefined where the consensus produced no
label (\cref{tab:appendix-human-strata}; method $\kappa=0.61$, direction
$\kappa=0.29$). Where the reference is
solid, humans and the consensus coincide; where it is weak, they often do not,
so cross-model variation in L4 partly measures reference-label noise rather
than model behaviour. Exp~A accordingly enters this paper as a paper-context
design-recognition diagnostic under source-derived contexts, and no
execution-level claim rests on it (\cref{sec:limitations}).
\begin{figure}[h]
\centering
\includegraphics[width=0.9\textwidth]{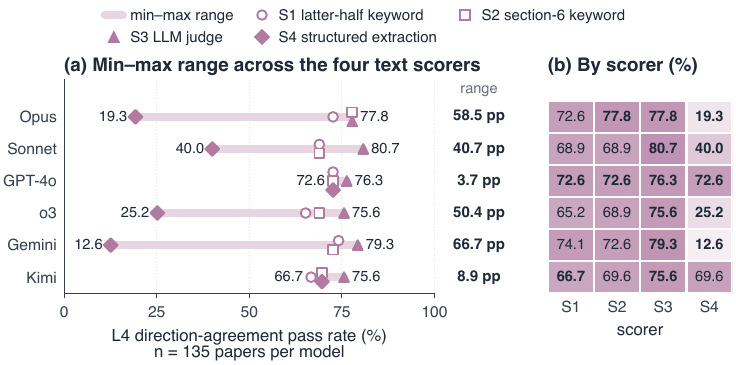}
\caption{L4 scorer instability across frozen scorer variants (ported from
the v1 release). Direction-agreement scores move materially under
equally-defensible scorer variants, supporting the label-noise bound on
text-layer diagnostics.}
\label{fig:l4-instability}
\end{figure}

\begin{figure}[h]
\centering
\includegraphics[width=0.90\textwidth]{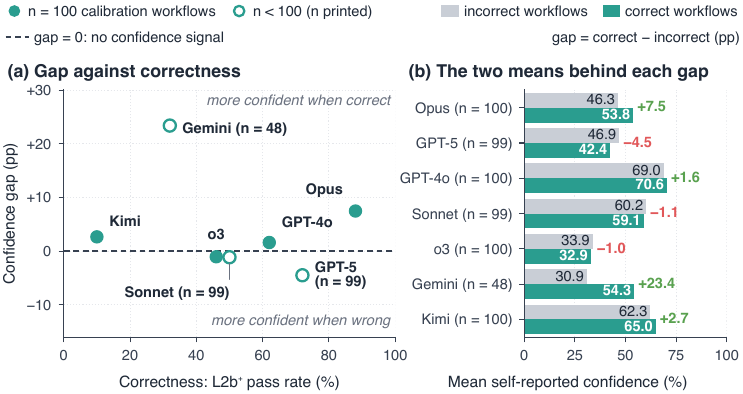}
\caption{Correctness versus self-assessment. \textbf{(a)} Each point is one
model: horizontal, execution-grounded L2b$^{+}$ correctness; vertical, the mean
confidence gap between correct and incorrect workflows. \textbf{(b)} The two
means behind each gap: mean self-reported confidence on incorrect (grey) and
correct (teal) workflows, with the signed gap. High correctness does not imply
a usable confidence signal.}
\label{fig:calibration-gap}
\end{figure}

\paragraph{Self-reported confidence does not triage.} Calibration asks each
model to rate its own prior Exp~B output without seeing the pass/fail label~\citep{kadavath2022know,lin2022uncertaintywords,tian2023calibration,xiong2024uncertainty}.
The elicitation asks for confidence that the estimate lies within 20\% of the
true effect, whereas correctness is scored at 50\% of the canonical estimate,
so the two are not the same event.
Actual correctness varies sharply across the panel (10.0\% to 88.0\%
L2b$^{+}$), yet the confidence gap (mean confidence on correct minus
incorrect outputs) stays within $\pm 0.075$ for every main-panel model;
the one larger gap (Gemini, $+0.234$) sits on the smallest usable sample
($n{=}48$). The asymmetry is the point: GPT-5 is second on execution-grounded
correctness with a slightly \emph{negative} gap, so ability and
self-assessment dissociate in both directions (\cref{fig:calibration-gap}).
Under this retrospective protocol, self-reported confidence is not a usable
deployment triage signal; the result does not rule out stronger calibration
interfaces, and is evidence against assuming one exists by default.

\paragraph{Cross-vendor robustness.} A single open-weights check
(Llama-3.3-70B-Instruct, same 100 scenarios, same pipeline) reproduces the
runnable-versus-correct gap: L2b $=41\%$ against final L2b$^{+}=20\%$,
placing it between Kimi and Gemini. Adding it as an eighth model lowers the
cross-layer rank correlation to Kendall $\tau=0.714$ without changing any
qualitative ordering. This is a robustness check, not a benchmark of the
open-weights ecosystem.

\section{Human Annotation and Coefficient-Extraction Audits}\label{sec:appendix-human}

Two blinded human audits bound the two instruments that are not executable
by construction: the consensus reference labels of the real-paper arm
(\cref{sec:expa}) and the coefficient-extraction judge of the executable arm
(\cref{sec:controls}). Both were run after the automated benchmark freeze,
and neither changes any frozen label, prompt, output, threshold, or headline
rate.

\subsection{Real-paper reference labels: 30-paper ambiguity audit}
\label{sec:appendix-human-labels}

\paragraph{Protocol.} The release contains a paper-native labeling protocol
and a completed, blinded 30-paper slice of the real-paper arm. A single
annotator labeled method family and effect direction from the original PDFs
without access to the LLM votes, the consensus labels, or any model output;
the consensus labels are the references against which \lmethod\ (legacy code
L3) and \ldirection\ (legacy code L4) are scored. Direction labels rest on three of the four
voters, since one voter casts no direction vote. One target paper
was excluded because the available PDF was appendix-only, and a blinded
replacement was drawn from the unsealed sample manifest, which contains only paper ID, difficulty
tier, and PDF path. The human labels are an ambiguity check, not a training,
prompting, or tuning input: they are not used to change the frozen consensus
labels, construct prompts, select model outputs, or tune scoring thresholds.
With one annotator and no relabel pass, every statistic below is agreement
between human and consensus, not inter-annotator reliability. The slice is a
stress test rather than a representative sample: 21 of its 30 papers
(70.0\%) lie in the low-consensus strata of the four-LLM vote (2-of-4
plurality, 2-of-4 tie, split), against 116 of 259 (44.8\%) corpus-wide.

\paragraph{Agreement and denominators.} Method-family labels match on 18 of
the 30 papers (60.0\%). Five of the 30 (three 2-of-4 ties, two splits) have
no consensus method label and count as non-matches in that figure; on the 25
papers where both labels exist agreement is 18/25 (72.0\%), and Cohen's
$\kappa = 0.606$ is computed on those 25. Nine papers have no consensus
direction label, leaving 21: direction agreement is 10/21 (47.6\%) with
$\kappa = 0.294$, where ``unclear'' is treated as a label (in 9 of the 11
non-matches the consensus says ``unclear'' and the human label does not).
The 60.0\% and 72.0\% figures are thus the same 18 matches over two
denominators. \Cref{tab:appendix-human-strata} gives the breakdown by
consensus strength: agreement is complete where at least three of the four
labelers agree and 56.2\% under a bare plurality, while the tie and split
rows are zero by construction, because there is no consensus label to match.
The audit supports reading the real-paper arm as an agreement diagnostic with
substantial ambiguity; it is not a full human re-adjudication of the
259-paper corpus.

\begin{table}[h]
\centering
\caption{Method-family agreement between the blinded human labels and the
four-LLM consensus, by strength of the consensus vote. Light-gray rows group
the strata by whether a consensus method label exists; the bold row is the
headline figure, and agreement values are left uncoloured because they are
not a ranking.}
\label{tab:appendix-human-strata}
\small
\setlength{\tabcolsep}{5pt}
\begin{tabular}{lcccc}
\toprule
Consensus stratum & Consensus label & Papers & Matches & Agreement \\
\midrule
\grouprow{5}{Consensus label exists}
4 of 4 agree & yes & 4 & 4 & 100.0\% \\
3 of 4 agree & yes & 5 & 5 & 100.0\% \\
2 of 4, plurality & yes & 16 & 9 & 56.2\% \\
\grouprow{5}{No consensus label (zero by construction)}
2 of 4, tie & none & 3 & 0 & 0.0\% \\
Split & none & 2 & 0 & 0.0\% \\
\midrule
\textbf{All audited papers} & & \textbf{30} & \textbf{18} & \textbf{60.0\%} \\
Papers with a consensus label & & 25 & 18 & 72.0\% \\
\bottomrule
\end{tabular}
\end{table}

\subsection{Coefficient-extraction judge: 50-cell audit}
\label{sec:appendix-human-coef}

\paragraph{Protocol.} The coefficient-extraction judge is a measurement
instrument: it identifies the scalar treatment-effect coefficient from
executed R code and its stdout before \lcorrect\ (legacy code L2b$^{+}$)
compares that coefficient to the canonical estimate. The annotation form of
the blinded audit hides model identity, the \lcorrect\ label, the canonical
estimate, the judge's and the legacy regex extractor's effects, and all
ranking information. The completed audit covers 50 v1-panel cells that pass
\lexec\ (legacy code L2b), sampled with seed 20260502: DID 13, Event Study
12, IV 11, RDD 14. Sampling prioritizes cells whose pass/fail label changed
between successive scorer generations (35 of the 50), so hard extraction
cases are over-represented.

\paragraph{Results and denominators.} All 50 cells were annotated. In 6 of
them (DID 1, IV 1, RDD 4) the annotator found no treatment-effect
coefficient in the output, so there is no human value to compare; in five of
these the judge also extracted nothing, and in one RDD cell it extracted a
value that passes. Both agreement rates are therefore out of the 44
comparable cells, not out of 50. Numeric agreement between judge and human
is 40/44 (90.9\%), where agreement means an absolute difference of at most
$10^{-4}$ or a relative difference of at most 1\%; agreement on the induced
\lcorrect\ pass/fail decision, applying the default 50\% tolerance to the
human-extracted value, is 39/44 (88.6\%). The annotator flagged 15 of the 50
cells (30.0\%) as having more than one plausible target coefficient. The
audit summary flags 12 cells in all (DID 1, Event Study 6, IV 1, RDD 4): the
six Event Study cells carry every disagreement among the 44 comparable cells
(all four numeric and all five pass/fail disagreements), and the other six
are the non-comparable cells. Most flags thus arise from two sources: (i)
Event Study outputs that report a cumulative post-event window rather than
the canonical per-day scale, and (ii) RDD outputs where printed signs, jump
orientation, or robustness estimates make the main effect ambiguous. The
audit supports the extraction instrument and does not change the frozen
headline \lcorrect\ rates; the flagged cases are why \lcorrect\ is described
as an audited measurement instrument rather than an infallible oracle.

\section{Extended Related Work}\label{sec:appendix-related}

Prior work evaluates important pieces of the LLM causal-analysis workflow,
but usually not the executed causal estimate~\citep{yang2024criticalreview,ma2025causalsurvey}. Causal benchmarks test textual
reasoning about direction, counterfactuals, graphs, interventions, or
identification~\citep{jin2024corr2cause,jin2023cladder,chen2024clear,chi2024unveiling,kiciman2023causal}. Code benchmarks test whether generated programs execute or
pass functional tests~\citep{chen2021codex,hendrycks2021apps,liu2023evalplus,zhuo2024bigcodebench,jain2024livecodebench}. Science-agent benchmarks evaluate broader discovery or
tool-use workflows. \textsc{CausalVerify} differs in its verification target:
the primary correctness signal is not a text label or a generic program test,
but whether executed model-written analysis code recovers a benchmark-defined
target estimate on the realised dataset.

Among end-to-end causal benchmarks,
CauSciBench~\citep{acharya2025causcibench} evaluates
causal-analysis performance over
curated real, synthetic, and textbook tasks, graded on the final product.
\textsc{CausalVerify} asks a complementary measurement question: when a
model appears to understand a causal task, \emph{which layer of the workflow
is externally verified}? It separates design-recognition diagnostics,
controlled-DGP execution verification, and retrospective calibration rather
than collapsing them into one causal-ability score. Its paired protocol
adds an instrument no adjacent benchmark carries: a design committed and
hashed before any data are revealed, then executed, so design-layer and
execution-layer failures separate on the same instance rather than across
task pools.

Concurrent work by \citet{sawarni2026causalreasoningbenchmark} disentangles causal
identification from estimation on real-world datasets; its finding that
strategy identification is easier than full specification is consistent with
our name-versus-compute gap, and our six-pair pilot sharpens the same
contrast to the single trajectory: a correctly named family whose
committed specification diverges and whose faithful execution is therefore
wrong on every seed. The methodological difference is the unit of
verification: \textsc{CausalVerify} verifies generated and executed code on
controlled DGP data, extracts the reported coefficient, and checks it
against a realised-data canonical estimate under frozen bands. This isolates
the failure mode where code runs but computes the wrong causal quantity.
Recent workflow-oriented evaluations such as
BioMysteryBench~\citep{anthropic2026biomysterybench} place models in rich
environments with real data, tool access, and objective final answers.
\textsc{CausalVerify} is complementary: it fixes a canonical estimate on
the realised data so that model-written causal code can be audited layer by
layer rather than graded only on a final answer.

Three adjacent lines of work appeared around and after the v1 release.
CausalDS~\citep{leban2026causalds} evaluates data-science agents across
Pearl's three rungs~\citep{pearl2009causality} on synthetically generated structural causal models, with
ground truth derived from the known structures; \textsc{CausalVerify}
occupies the narrower estimation slice at higher resolution: paper-grounded
twins, canonical estimators under frozen tolerance bands, and a
commitment-before-data protocol that separates the design layer on the same
instance. InterveneBench~\citep{shi2026intervenebench} scales design-layer
evaluation to 744 real policy studies, asking whether models can construct
an appropriate causal study design from open-ended intervention contexts.
Its STRIDES system uses simulated data and code execution internally to
refine proposed designs, but the benchmark endpoint remains agreement
between the predicted and expert-verified study design rather than
numerical recovery of an externally fixed causal estimate.
\textsc{CausalVerify} addresses the complementary downstream verification
problem: once a model commits to a design, did it implement that design
faithfully, and did the execution recover the benchmark-defined estimand?
The two evaluations are end-to-end over different chains (study-design
generation there, verification here) with different terminal endpoints. Closest in verification style, \citet{kohler2026readthepaper} reproduce social-science results from
paper text and original data with deterministic cell-level comparison to the
published outputs; the criterion there is agreement with the original
authors, whereas \textsc{CausalVerify}'s canonical estimates are
benchmark-fixed on synthetic twins, so recovering the target is separable
from matching a possibly imperfect original. Recitation of published
values becomes a threat its referee zones must control rather than the
success criterion. The oracle arm is likewise not a reproduction task: it
substitutes the gold design against the model's \emph{own prior commitment}
on the same instance, with a twin-fixed target. ReplicatorBench~\citep{nguyen2026replicatorbench} grades
agents across the stages of a replication pipeline against human-verified
replicable and non-replicable claims. It evaluates competence \emph{at}
replication; \textsc{CausalVerify} fixes the design layer via
committed CDRs and verifies the executed estimate itself.
\Cref{tab:benchmark-comparison} summarizes the affordances.

\begin{table}[h]
\centering
\caption{Comparison with adjacent benchmarks. The table reports evaluation
affordances rather than overall benchmark quality: \Lpass\ marks an explicit
mechanism for that dimension, not superiority; \Lpart~= partial; \Lfail~=
absent. $^{\dagger}$EconCausal counts 10{,}490 triplets across 2{,}595
studies.}
\label{tab:benchmark-comparison}
\footnotesize
\renewcommand{\arraystretch}{1.08}
\setlength{\tabcolsep}{2.5pt}
\resizebox{\textwidth}{!}{%
\begin{tabular}{lclcccc}
\toprule
\textbf{Benchmark} & \textbf{N} & \textbf{Target} &
\textbf{Real ctx.} & \textbf{Exec.\ coeff.} &
\textbf{Layered Dx} & \textbf{Self-assess.} \\
\midrule
CLadder / CORR2CAUSE~\citep{jin2023cladder,jin2024corr2cause} & 11k / 400 & Causal QA / direction
  & \Lpart & \Lfail & \Lfail & \Lfail \\
CausalBN-Bench~\citep{zhou2024causalbench} & 3 & Graph + effect
  & \Lfail & \Lfail & \Lpart & \Lfail \\
EconCausal~\citep{lee2025econcausal} & 10{,}490$^{\dagger}$ & Economic causal sign
  & \Lpass & \Lfail & \Lfail & \Lfail \\
CauSciBench~\citep{acharya2025causcibench} & 367 & End-to-end CI pipeline
  & \Lpass & \Lpart & \Lpart & \Lfail \\
InterveneBench~\citep{shi2026intervenebench} & 744 & Intervention design
  & \Lpass & \Lpart & \Lpart & \Lfail \\
CausalReasoningBench~\citep{sawarni2026causalreasoningbenchmark} & 173 & ID + estimate
  & \Lpass & \Lfail & \Lpart & \Lfail \\
Code benchmarks~\citep{chen2021codex,jimenez2024swe} & -- & Program tests
  & \Lfail & \Lpart & \Lpart & \Lfail \\
Discovery / science agents~\citep{majumder2024discoverybench,chen2024scienceagentbench} & 264 / 102 & Discovery workflow
  & \Lpass & \Lpart & \Lpart & \Lfail \\
\textsc{CausalVerify} & 259 + 100 + 23 & Causal workflow
  & \Lpass & \Lpass & \Lpass & \Lpass \\
\bottomrule
\end{tabular}}
\end{table}

\section{Benchmark Controls}\label{sec:appendix-controls}

\begin{table}[h]
\centering
\caption{Benchmark controls. Each component supports a different claim and
protects against a different failure mode.}
\label{tab:controls}
\small
\renewcommand{\arraystretch}{1.12}
\setlength{\tabcolsep}{3pt}
\begin{tabularx}{\textwidth}{p{0.18\textwidth}p{0.40\textwidth}X}
\toprule
\textbf{Component} & \textbf{Implementation} & \textbf{Claim supported} \\
\midrule
Real context & 259 papers; reconstructed RQ/DD/IC; consensus reference
labels & Text-level design-recognition diagnostics \\
Executable truth & 100 fixed-seed DGPs and 23 paper-grounded twins;
realised data; canonical estimators & Numerical coefficient correctness \\
CDR commitment & Stage-1 design hashed before any data are revealed; the
main arm executes exactly the committed design & Same-instance separation of
design from execution; no post-hoc design revision \\
Role separation & Separate stages for construction, response, labeling,
extraction, calibration & Reduces circular grading \\
Leakage control & Method-name blacklist; prompt-surface leakage gates;
opaque task identifiers; sealed truths & Prevents trivial label recovery \\
Scorer audit & Coefficient-extraction judge plus blinded human validation &
Reduces brittle post-processing errors \\
Contract freeze & Hashed twin-set freeze; dated amendments that state their
triggers, including when they are not result-blind & Distinguishes
pre-specified structure from post-hoc repair \\
Claim boundaries & Exp~A = recognition diagnostics; Exp~B = execution;
paired = same-instance cross-layer contrast; calibration = confidence alignment &
Prevents overclaiming \\
\bottomrule
\end{tabularx}
\end{table}

\section{Limitations: Extended Discussion}\label{sec:appendix-limitations}

\paragraph{Scope of the causal task.} The benchmark evaluates structured
econometric estimation workflows after the research question, the data and the
institutional setting have been fixed. It does not evaluate problem
formulation, data acquisition, design selection from ambiguous field
constraints, or the broader causal-inference lifecycle in the sense of
\citet{hernan2019comment}; a model could excel here and still fail at the
stages this benchmark deliberately holds constant.

\paragraph{Real-paper labels are diagnostics, not ground truth.} Exp~A's
labels come from a four-LLM consensus whose pool structurally overlaps the
evaluated panel. The blinded 30-paper ambiguity audit agrees with that
consensus on 60.0\% of method labels and 47.6\% of direction labels
(direction $\kappa=0.294$). Exp~A therefore supports design-recognition
diagnostics under source-derived paper contexts; it is not evidence about
unaided design inference from neutral institutional facts, and no
execution-level claim rests on it. A separate construct audit of the legacy
Exp~A contexts remains open and is tracked in the released audit records.

\paragraph{Synthetic structure and backend.} Executable reference estimates
require synthetic twins; the DGPs cover four design families, not the space of
applied practice. The v1 corpus fixes R as the execution backend while the
paired arm executes Python, so cross-arm comparisons carry a backend
difference as well as a protocol difference; v1's Python-backend replication
shows the executed-but-wrong phenomenon is not R-specific, but magnitudes
should not be read as backend-general.

\paragraph{Commitment before data.} Stage~1 commitments close before the data
dictionary is visible, and the single-shot contract offers no renegotiation. A
commitment can therefore be faithful to the source literature and still
diverge from the benchmark gold and fail to execute as committed against the
twin's shipped geometry (\cref{sec:pilot}). The protocol records this as a
main-arm outcome by design; the one-round repair pilot recovers
non-executing workflows more often than executed-but-wrong ones (33.0\%
versus 22.2\%).

\paragraph{Panel scale and model coverage.} The paired evidence now spans
the full 23-pair, nine-model panel (\cref{sec:panel}); the six-pair pilot
(\cref{sec:pilot}) remains as harness validation and taxonomy. Conditional
downstream rates are reported with pair-clustered uncertainty, and the
recovery-difference interval includes zero. The evaluated panels, v1 and panel,
consist mainly of closed or hosted systems; rank statements are descriptive
over the evaluated panel, not population claims over all models.

\paragraph{The instrument is audited, not infallible.} Scoring depends on
canonical specifications, frozen tolerances and coefficient extraction;
individual-cell mistakes remain part of measurement uncertainty. Changes made
after evidence existed are recorded as dated amendments that state their
trigger, including two that are explicitly not result-blind (a unit-contract
clarification and the repair of one task's Stage-1 context after the pilot),
alongside a design-profile instrument that was itself built after pilot
outcomes were known, so a reader can
distinguish pre-specified structure from post-hoc repair rather than being
asked to assume there was none.

\paragraph{Contamination and decay.} The paired release re-keys every
model-facing task to opaque identifiers; method family and difficulty appear
in no filename, identifier or model-visible record, ground truth lives
scorer-side, and per-twin truths remain sealed. Because source papers are
published, familiarity from pretraining cannot be fully excluded on the
real-paper side. Any released benchmark decays as later models train on its
artifacts~\citep{kiela2021dynabench,jain2024livecodebench}; released results are versioned against hashed freezes, and sealed
holdout material supports future re-keyed refreshes.

\section{Additional Panel and Legacy Diagnostics}\label{sec:appendix-panel}

\begin{figure}[h]
\centering
\includegraphics[width=0.85\textwidth]{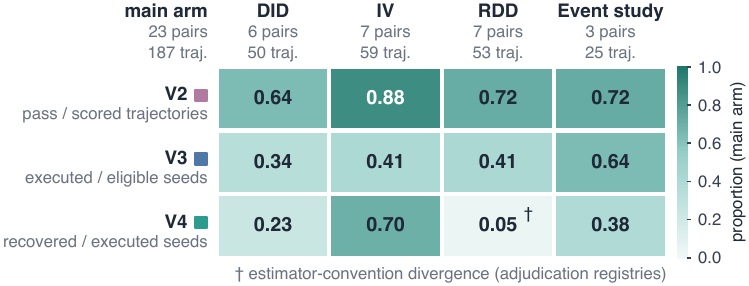}
\caption{Panel profile by causal-design family (main arm, descriptive).
Each layer keeps its own denominator; pairs per family are few, so no
family ranking is implied. The RDD cell's low recovery is dominated by
estimator-convention divergence documented in the adjudication registries.}
\label{fig:panel-family}
\end{figure}

\begin{figure}[h]
\centering
\includegraphics[width=0.8\textwidth]{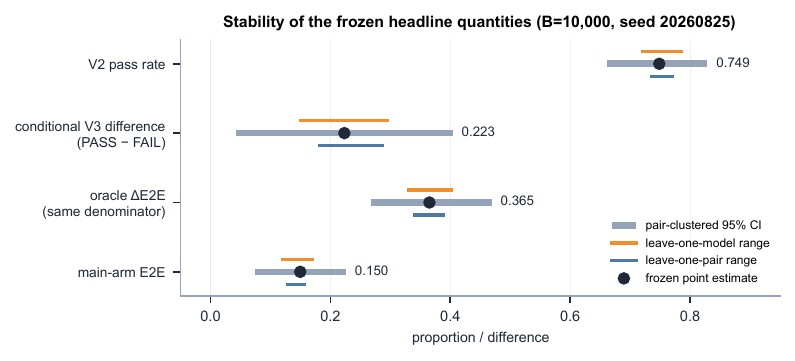}
\caption{Stability of the frozen headline quantities: pair-clustered 95\%
bootstrap CIs with leave-one-model and leave-one-pair ranges. The
substitution headline moves by at most $\pm 3.7$pp under any leave-one-out;
the other quantities move by up to 7.9pp.}
\label{fig:panel-loo}
\end{figure}

\begin{figure}[h]
\centering
\includegraphics[width=0.9\textwidth]{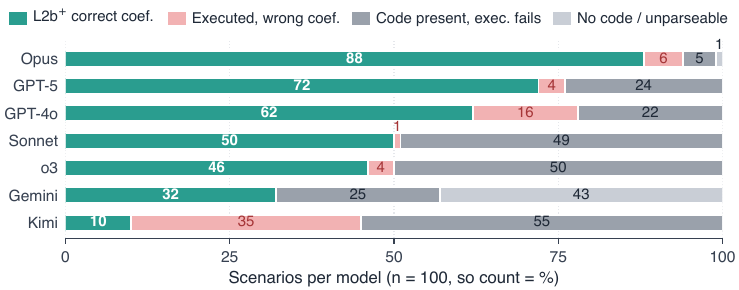}
\caption{Legacy execution cascade from the v1 100-scenario DGP panel
(demoted from the v1 main text): per-model composition of L2b$^{+}$
outcomes. Retained for continuity with the v1 release; the paired panel of
\cref{sec:panel} supersedes it as the paper's core evidence.}
\label{fig:legacy-cascade}
\end{figure}

\paragraph{$\geq$1-seed end-to-end figure (descriptive).} 31 of 187 scored
trajectories (15.0\% of the 207 attempts) achieve Stage-1 validity, a V2
pass, at least one executed seed and at least one jointly recovered seed.
The threshold was not pre-frozen; the continuous joint-recovery proportion
over eligible seeds in \cref{sec:panel} is the primary end-to-end metric.

\end{document}